\documentclass[lettersize,journal]{IEEEtran}
\usepackage{times}
\usepackage{graphicx}
\usepackage{amssymb}
\usepackage{url,hyperref}
\hypersetup{hidelinks,
  pdftitle={A Structure-Agnostic Multi-Objective Evolutionary Weight-Sharing Compression Framework for Deep Neural Networks},
  pdfauthor={Rasa Khosrowshahli, Shahryar Rahnamayan, Beatrice Ombuki-Berman}}
\usepackage{booktabs}
\usepackage{multirow}
\usepackage{subcaption}
\usepackage{placeins}
\usepackage{float}
\usepackage{algorithm,algpseudocode}
\usepackage{amsmath}
\usepackage{adjustbox}
\usepackage{mathtools}
\makeatletter
\renewcommand{\eqref}[1]{\textup{Eq.~\tagform@{\ref{#1}}}}
\makeatother

\usepackage{lipsum} 
\usepackage{textcomp}
\usepackage{stfloats}
\usepackage{verbatim}
\usepackage{graphicx}
\usepackage{cite}
\usepackage{xcolor}
\usepackage{tikz}
\usetikzlibrary{arrows.meta,calc,decorations.pathreplacing}

\begin{document}
\bstctlcite{IEEEexample:BSTcontrol}

\title{A Structure-Agnostic Multi-Objective Evolutionary Weight-Sharing Compression Framework \\ for Deep Neural Networks}

\author{Rasa~Khosrowshahli,~\IEEEmembership{Graduate Student Member,~IEEE,}
        Shahryar~Rahnamayan,~\IEEEmembership{Senior Member,~IEEE,}
        and~Beatrice~Ombuki-Berman,~\IEEEmembership{Member,~IEEE}%
\thanks{This work was supported by the Natural Sciences and Engineering Research Council of Canada (NSERC) through a Discovery Grant. \textit{(Corresponding author: Shahryar Rahnamayan.)}}%
\thanks{Rasa Khosrowshahli and Beatrice Ombuki-Berman are with the Department of Computer Science, Faculty of Mathematics and Science, Brock University, St.~Catharines, ON L2S~3A1, Canada (e-mail: rkhosrowshahli@brocku.ca; bombuki@brocku.ca).}%
\thanks{Shahryar Rahnamayan is with the Department of Engineering, Faculty of Mathematics and Science, Brock University, St.~Catharines, ON L2S~3A1, Canada (e-mail: srahnamayan@brocku.ca).}%
\thanks{\copyright~2026 IEEE. Personal use of this material is permitted. Permission from IEEE must be obtained for all other uses, in any current or future media, including reprinting/republishing this material for advertising or promotional purposes, creating new collective works, for resale or redistribution to servers or lists, or reuse of any copyrighted component of this work in other works. Accepted for publication in \textit{IEEE Transactions on Big Data}. Digital Object Identifier 10.1109/TBDATA.2026.3744883}}
\markboth{IEEE Transactions on Big Data}{IEEE Transactions on Big Data}

\maketitle

\begin{abstract}
Deep neural networks (DNNs) have large parameter sets that are costly to store and to deploy on resource-constrained devices. Weight-sharing compression replaces the many distinct weight values of a network with a small codebook of shared values, called codewords. Prior methods build the codebook by clustering, typically with K-means, which is expensive on models with millions of parameters and must be refit for every codebook size. To stay tractable, clustering is often applied per layer, which requires the network's layer boundaries, tensor shapes, and channel counts.
We propose Uniform-Binned Entropy-Constrained Scalar Quantization (UB-ECSQ) with NSGA-II, a retraining- and gradient-free, structure-agnostic, multi-objective compression framework that treats the network as a closed-box. Rather than compressing layer by layer, it flattens all parameters into a single vector that shares one codebook. Uniform Binning Quantization (UBQ) maps the vector onto equal-width uniform bins in linear time and represents each occupied bin by the centroid of the values in that bin. Entropy-Constrained Scalar Quantization (ECSQ) then merges neighboring bins, trading the added distortion against the entropy of bin occupancy. With the ordered bins, we use dynamic programming to find the optimal merge using only bin statistics, without any forward or backward pass. We formulate compression as a multi-objective optimization problem with two objectives and use NSGA-II to search over only two variables, the number of bins and the merge ratio, and it returns a Pareto front of compressed models that trade accuracy against compression ratio (CR).
Within a 1\% drop in accuracy, and averaged over five seeds, our Pareto fronts reach CRs of 20.8 and 12.9 for ResNet-18 on CIFAR-10 and CIFAR-100, 10.4 to 12.3 for ResNet-18/34/50/101 on ImageNet-1K, 18.4 for VGG16-BN, and 9.0 for ViT-B/16. At this budget, our framework attains a higher CR than every baseline evaluated in each setting. Over the whole objective space, our final Pareto front also achieves the highest hypervolume in all eight settings.
\end{abstract}


\section{Introduction}
\IEEEPARstart{T}{he} demand for compressing leading convolutional networks (CNNs) is increasing to enable their application in resource-constrained devices such as self-driving autonomous vehicles and mobile assistant robots \cite{ota2017deep}.
Storing the set of parameters known as weights for AI models in memories is challenging \cite{han2015deep}.
Over the past decade, ConvNets have been steadily developed and improved to generalize in object detection and segmentation since AlexNet \cite{krizhevsky2012imagenet} in 2012, then ResNet \cite{he2016deep} in 2016, and Vision Transformers (ViT) \cite{dosovitskiy2020image} in 2020. These models are now extremely complex in capturing and extracting features accurately, leading to large networks with millions of parameters.
Recent years have seen a lot of research focused on compressing the deep neural networks (DNNs), which has produced networks that fit in 1 MB on ImageNet-1K classification \cite{wang2019haq, stock2019and}.
These compression settings are also highly relevant to real-time and edge deployment scenarios, such as autonomous systems, mobile robotics, embedded systems, and TinyML applications, where compact model storage and low-complexity processing are critical \cite{sze2017efficient,lin2020mcunet}.

However, some applications, such as multi-task learning, require merging or combining ConvNets or language models \cite{chou2018merging, goddard2024arcee}. To optimize a coupled network like encoder-decoder architecture, it is suggested to adopt a hard or soft parameter-sharing technique that reduces the active parameters \cite{duong2015low,ruder2017overview}.
Hard parameter sharing is commonly suggested to have multiple task-specific layers and one shared layer that contains the shared parameters \cite{caruana1997multitask,ruder2017overview}.
In general, sharing parameters not only reduces memory but also eases optimization \cite{caruana1997multitask}.

One can ask if we really need $D$ quantities of variables to describe one object. The intrinsic dimensionality of a space refers to the required quantity of information elements needed to formally characterize each object within the space \cite{li2018measuring}.
For instance, in signal processing of multi-dimensional signals, the intrinsic dimension of a signal represents the necessary variables to generate a good approximation of the signal \cite{li2018measuring}. 
In contrast, the number of quantities we use is called extrinsic $D$-dimensionality of the space \cite{li2018measuring}. The intrinsic dimensionality of space is smaller than its extrinsic dimensionality when the subspace is a proper subset of the full extrinsic space. This concludes with the essence of a new, smaller, and more efficient space than the full $D$-dimensional space, which is called subspace. 

In the case of neural networks, let us consider a network with $N$ total weights as parameters and let $ \theta \in \mathbb{R}^N$ be a parameter vector in a parameter space of dimension $N$. Here the ambient extrinsic dimension coincides with the parameter count, so we identify $D$ with $N$ whenever we refer to the full parameter vector. It is possible to envision the objective landscape as a collection of ``hills and valleys'' in $N$ dimensions, with a value of the loss associated with each point in $\mathbb{R}^N$ \cite{li2018measuring}. 
One solution is to reduce the dimensionality of the parameter space from the full extrinsic dimension $N$ to intrinsic dimensions $D_{\mathrm{int}}$, but it must not change the location of the parameter vector in the objective landscape, known as performance degradation \cite{li2018measuring}. 

Various methods are studied for compressing neural networks, such as quantization, pruning, low-rank approximation, and knowledge distillation, which are the most popular approximation techniques \cite{rokh2023comprehensive}. The primary focus is to find the intrinsic weights with the least possible performance loss. The main difference is maintaining or cutting the connections between neurons in neural networks.

In particular, pruning is used to trim the least important connections of the network by zeroing the weight value, which reduces the number of non-zero parameters and increases network efficiency \cite{han2015learning,liang2021pruning}.
However, pruning without retraining significantly degrades accuracy \cite{han2015learning}.

Quantization is a widely studied area that explores various methods for reducing the precision of numerical representations in neural networks. It involves minimizing the precision of values used for weights, biases, and/or activations. The primary objective is to decrease the model's computational and memory demands while preserving a satisfactory level of accuracy \cite{jacob2018quantization,rokh2023comprehensive}.

There are two types of quantization that aim to quantize networks, but in distinct approaches described as follows.
The first type is bit-precision quantization, which focuses on reducing weight precision by mapping weights to a discrete range based on bit width \cite{jacob2018quantization}. In general, weights are stored and represented in float 32 bits equally, but it is not essential to maintain high resolution and accurate values to build up the network where it performs the same.
The second approach focuses on representing the network's weights with significantly fewer unique values while maintaining its performance. This is accomplished by reducing weight diversity through clustering techniques \cite{han2015deep,gong2014compressing}. Clustering, an unsupervised method, identifies groups of similar values. Similarity within the weight space is determined by minimizing the distance between weights and cluster centroid.

Han \textit{et al.} \cite{han2015deep} proposed a weight-sharing quantization method using clustering techniques such as K-means for weight quantization, where each weight belonging to the same cluster is represented by a shared centroid weight. 
These clustering techniques use distance metrics to find the representative centroid and group of closest and similar weights based on their values. They are limited to finding the golden $K$ parameter, which determines the a priori number of codewords. That said, grouping the weights using the K-means algorithm has $\mathcal{O}(N~K~T~R)$ complexity, where $T$ is the number of iterations to optimize and $R$ is the number of random restarts, making it costly to optimize for millions of parameters $N$ in DNNs \cite{lloyd1982least}. 
Prior work therefore swept $K$ over powers of two, because a fixed-width index uses $\lceil\log_2 K\rceil$ bits on conventional memory hardware and because the cost of K-means made intermediate $K$ values between consecutive exponents impractical to evaluate \cite{han2015deep}.

To keep K-means tractable, some works apply it per layer, splitting the total cost across smaller blocks \cite{han2015deep,wu2016quantized,dupuis2022heuristic}. Bit-precision methods similarly adopt per-layer or per-channel scales \cite{krishnamoorthi2018quantizing,jacob2018quantization}. That decomposition, however, requires knowledge of the network's internal structure, such as layer boundaries, tensor shapes, and channel counts, which is unavailable when the model is released as a closed-box or closed-source artifact.

Most previous works used distance-based clustering algorithms with the representative centroid value and achieved good performance with pruning \cite{han2015deep, wu2018deep, son2018clustering, wu2016quantized, gong2014compressing}. The resulting clusters and cluster centroid values are not the optimal form of quantization, leading to accuracy loss reported by Pikoulis \textit{et al.} \cite{pikoulis2022new}.
The majority of these methods and \cite{ullrich2017soft} recommended retraining the constrained network with a small set of weights in order to recover the performance of the compressed network. Retraining, however, can be expensive and resource-intensive.
Nonetheless, there has to be an efficient and effective way to cluster and share representative weights without requiring retraining.
In order to avoid retraining, Dupuis \textit{et al.} \cite{dupuis2021cnn, dupuis2022heuristic} proposed retraining-free weight-sharing for network compression approach using heuristics on ImageNet-1K dataset.

Studying in DNNs, the entries $\theta_j$, $j = 1, \ldots, N$, of the weight vector $\theta \in \mathbb{R}^N$ are initialized independently from a zero-mean normal distribution \cite{he2015delving}, and the network is trained on normalized data inputs. 
After training, the empirical histogram of the weights is typically bell-shaped and concentrated near zero \cite{han2015deep,park2017weighted,banner2019post}.
More accurate quantization can be achieved if the distribution of quantization levels is in line with that histogram \cite{park2017weighted,banner2019post}. 
Uniform quantization methods approximate these concentrated weights by mapping them to discrete levels on a uniform grid \cite{gish1968asymptotically}. In the classical scalar-quantizer construction, the reconstruction value of each equal-width interval is a \emph{fixed} point that does not depend on the data (typically the geometric midpoint), so empty intervals still occupy a level. They are usually used for bit-precision quantization \cite{jacob2018quantization}. In comparison with non-uniform quantization, a uniform grid is more efficient but less precise in clustering quality.

Uniform quantization methods are simpler and faster at large scale \cite{jacob2018quantization,krishnamoorthi2018quantizing}, and they can be applied to the full parameter vector without inspecting architecture. Equal-width binning is a single pass over the $N$ weights, so a wide range of candidate bin counts can be explored without refitting a clustering procedure. Uniform Binning Quantization (UBQ), the first stage of our pipeline, uses the same equal-width cuts but a different reconstruction. Its stored codeword is the empirical mean of the weights that fell in the bin, and empty bins are dropped, so the levels sit on the data rather than on the grid centers.

Although the compression and quantization focus on finding optimal groups of weights, the primary question of ``\textit{how many codewords?}'' is still not satisfied, which has a major impact on compression rate and performance. Instead of relying on grid search, we aim to determine the optimal binning and merging configuration by leveraging evolutionary multi-objective algorithms like NSGA-II \cite{deb2002fast}, which are designed to simultaneously optimize two (and more) objectives where we set one objective as the Huffman-coded compression ratio and the other as accuracy on a calibration split.

Uniform bins, however, are not the final set of codewords. Neighboring bins that are similar, or that hold very few weights, can be merged to shrink the number of codewords further and, more importantly, to reshape the occupancy distribution so that Huffman coding of the index stream is cheaper. Because the stored model uses variable-length (Huffman) codes rather than fixed-width indices, the true storage cost depends on the entropy of bin occupancy, not merely on the number of bins. We therefore merge neighboring bins by Entropy-Constrained Scalar Quantization (ECSQ) \cite{chou1989entropy} which is an exact dynamic programming merge of neighboring bins, computed from bin statistics alone, that trades extra distortion against occupancy entropy inside every evolutionary evaluation.

Unlike prior pipelines that prune connections before or alongside quantization and then fine-tune to recover accuracy \cite{han2015deep,tung2018clip}, our method neither prunes nor retrains. It re-expresses a pretrained parameter vector in a smaller alphabet. Pruning is orthogonal to, and composable with, this pipeline. A sparse network is still a vector of real numbers, and the subsequent merge and Huffman stages apply unchanged to the surviving weights, so headline compression ratios that combine the two effects are not directly comparable to ours. Combining them is a natural extension that we leave to future work rather than claiming a combined compression ratio that we have not measured.

The main contributions of this work are summarized as follows:
\begin{itemize}
    \item We propose Uniform-Binned Entropy-Constrained Scalar Quantization (UB-ECSQ) with NSGA-II, a structure-independent weight-sharing quantization framework that compresses pretrained parameters with UBQ, without per-layer design or retraining of shared weights. The method treats all network parameters as a single one-dimensional vector, regardless of layer types, and the formulation does not inspect the architecture. We demonstrate it on convolutional and transformer image classifiers. We do not evaluate language models or non-DNN parametric models in this work and make no empirical claim about them.
    \item We formulate codeword construction as ECSQ. After UBQ, neighboring bins are merged by an exact dynamic program that minimizes extra distortion plus a weighted occupancy entropy, that is, the rate that Huffman coding of the index stream actually incurs. The inner distortion-rate trade-off is a classical Lagrangian scalarization.
    \item We place UBQ and ECSQ merging inside a closed-box multi-objective search over the initial bin count and the merge ratio, jointly minimizing accuracy loss and the reciprocal of the compression ratio. Merging is performed inside each evaluation, so the Pareto front already contains merged codewords suited to Huffman coding, and a decision maker can select a point that matches the target application's constraints. No merge step is applied after the search.
\end{itemize}

\indent{\raggedright The source code is publicly available at \url{https://github.com/rkhosrowshahi/EvoDeepQuantization}.\par}

The remainder of this paper is organized as follows: Section \ref{sec:related_works} reviews previous related works. Section \ref{sec:proposed_method} gives details of the proposed multi-objective, entropy-constrained compression of deep neural networks. Section \ref{sec:experiments} shows and analyzes experiments conducted on three well-known benchmark datasets on various types of DNNs. Lastly, this paper is concluded in Section \ref{sec:conclusion}.

\section{Related Works}\label{sec:related_works}
In this section, we provide the background and summarize related works on the compression of neural networks. We review quantization for storage, then work that treats the compressed parameter vector as a reduced search space in which fine-tuning can run. 

\subsection{Weight-Sharing Quantization}
Quantization of network parameters is either scalar or vector. Scalar quantization maps each weight independently to a reconstruction level on the real line. Reduced bit width on a uniform grid and one-dimensional weight sharing (a codebook of scalar centroids) are both scalar. Vector quantization maps a block of weights jointly to a codebook vector. Product quantization (PQ) is the vector method most often applied to neural network weights. This paper studies scalar weight-sharing. We return to PQ after that lineage, as a contrast.

Scalar weight-sharing aims to group $N$ weights into $K$ codewords ($K$ $\ll N$) and replace values with a shared weight. Thus, it allows the storing of only $K$ 32-bit float shared weights instead of $N$ 32-bit float weights. Usually, a lookup table (i.e., codebook) is constructed to map the shared representative weights to the associated weights in the neural network. Each codeword is a shared representative for $\mathcal{C}_i$ set of weights.
The rate of compression (CR) is calculated by using the following equation:
\begin{equation}
    CR = \frac{N \times b_{W}}{K \times  b_{W} + \lceil\log_2{K}\rceil \times \sum_{i=1}^{K} |\mathcal{C}_{i}| }
    \label{eq:cr}
\end{equation}
where $N$ is the total number of weights in the network, $K$ is the number of codewords (keys in the codebook), and $b_W$ is the number of bits used to represent weight values after training, which is usually 32 floating-point. $\mathcal{C}_{i}$ is the set of weights assigned to codeword $i$, with occupancy $|\mathcal{C}_i|$, where $\sum_{i=1}^{K} |\mathcal{C}_i| = N$. The ceiling function $\lceil\log_2{K}\rceil$ in \eqref{eq:cr} defines the number of bits required to represent the indices of the codewords.

Han \textit{et al.} \cite{han2015deep} proposed weight-sharing quantization using K-means clustering to construct a codebook of weight addresses, and to further increase the compression ratio, they applied Huffman coding. Their work consists of three compression techniques: 1) pruning connections (weights) by iteratively training the network, 2) quantizing by K-means clustering weights and generating a codebook, and 3) Huffman encoding weights and the representation indices. However, they retrained the pruned network with the quantized centroid weights. They showed that Huffman coding saves 20\%-30\% of memory. In this direction, the $CR$ \eqref{eq:cr} is modified by changing fixed-length bit indexing to variable-length indexing as follows:
\begin{equation}
    CR = \frac{N \times b_{W}}{K \times  b_{W} +  \sum_{i=1}^{K} b_{\mathcal{C}_i} \times |\mathcal{C}_{i}| }
    \label{eq:cr_huffman}
\end{equation}
where $b_{\mathcal{C}_i}$ is the Huffman bit length of the index of codeword $i$ in the compressed encoded codebook.

Consequently, the storage cost of the index stream is governed by the occupancy distribution of the codeword groups, not by the number of codewords alone. A codebook with many codewords but a highly skewed occupancy can compress better under Huffman coding \cite{huffman1952method} than a smaller codebook with a nearly uniform occupancy. Entropy-constrained vector and scalar quantization \cite{chou1989entropy,choi2020universal} exploit exactly this fact by trading distortion against rate rather than against the number of codewords.

Choi \textit{et al.} \cite{choi2016towards} proposed Hessian-weighted K-means clustering to quantize the network parameters after pruning. Inspired by uniform quantization \cite{gish1968asymptotically}, they uniformly spaced thresholds and divided network parameters into clusters. A classical uniform scalar quantizer reconstructs each interval by its midpoint. Their Uniform Quantization uses the per-bin mean of the weights that fall in the interval and therefore already differs from a midpoint grid and their Hessian-weighted variant replaces those means by Hessian-weighted means. However, Hessian computation needs to evaluate the second-order partial derivative of the differentiable loss function, which is a costly operation.

Distance-based clustering methods may not effectively capture or leverage the statistical properties of neural network weight distributions, especially if they deviate from the clustering method's assumptions.

There are also works that use different clustering techniques, such as DP-net \cite{yang2020dp}, which suggests using the dynamic programming technique rather than the K-means algorithm. Although they use a fixed and uniform number of shared values for each layer, they provide performance results after retraining.

Pikoulis \textit{et al.} \cite{pikoulis2022new} suggested a weight clustering method based on dictionary learning that lowers the overall computational complexity by giving the centroids a unique structure.

Ullrich \textit{et al.} \cite{ullrich2017soft} employed soft weight-sharing, a neural network regularization technique originally introduced by Nowlan and Hinton \cite{Simplifying}, to pretrain neural networks. Their method distributed 16 Gaussian mixture components uniformly across the range of pretrained weights, aiming to model the weight distribution through variational learning. However, this approach is computationally intensive and susceptible to network collapse if the Gamma hyperprior parameter for the variances of the mixture components is not included.

Dupuis \textit{et al.} \cite{dupuis2022heuristic} proposed a weight-sharing optimization heuristic approach using NSGA-II. They focused on bi-level optimization, where the first level is finding optimal $K$ codewords using K-means for each layer of the network with $l$ layers, and the second level is finding the optimal combination of each layer's $K$s based on accuracy loss prediction and $CR$.

In the first level, they search for an optimal set of $K$s in a small range for one layer, irrespective of considering the crucial impact of quantization in other layers on the whole network. As the second level is designed to find an optimal combination of $K$s, this kind of design space exploration can be ineffective, resulting in a non-global optimality solution. Considering a network with $l$ layers and $N$ weights, the cost of the second optimization level is expensive, as $\mathcal{O}\!\left(\prod_{i=1}^{l} |\phi_{i}|\right)$ where $\phi_{i}$ is the set of optimal $K$s from the first optimization level for layer $i$. Thus, they analyzed the layers of networks and cluster weights in layers separately to reduce the overjumping of shared weights across the network. Having a combination of various $K$s results in different numbers of codewords for each layer. However, they did not clarify the total number of codewords across multiple codebooks in their $CR$ equation. In addition, one can understand that using different $K$s for layers in the network requires the construction of multiple codebooks. For example, consider a network with $l=5$ layers where its layer's optimal $K$s are $K=\{5, 10, 18, 8, 200\}$, the minimum and maximum number of bits required to store indices of codebooks are $b^{\min}=\lceil \log_2 5 \rceil = 3$ and $b^{\max}=\lceil \log_2 200 \rceil = 8$. Although another approach is to use a single codebook, the maximum number of bits required to store indices $b^{\max}$ will be used.

As DNNs get deeper and more complex, compressing parameters of all layers together is even more advantageous since we can avoid optimizing the compression rate layer by layer \cite{choi2016towards}. Jointly optimizing the compression ratios across all individual layers incurs a time complexity that grows exponentially with the number of layers. This is because the joint search space is the Cartesian product of the compression ratio choices per layer, so its size, and therefore the cost of an exhaustive joint search, grows exponentially with the number of layers, with each additional layer multiplying the size of the combined search space by its own number of candidate ratios.

Product quantization (PQ) was initially proposed by Gong \textit{et al.} \cite{gong2014compressing} on fully connected (FC) layers, which occupy the most memory in ConvNets. Later, Wu \textit{et al.} \cite{wu2016quantized} explored PQ to compress both FC and convolutional layers of ConvNets. However, they observed that minimizing the quantization error is correctly reached by sequentially quantizing layers \cite{wu2016quantized}.

Stock \textit{et al.} \cite{stock2019and} use PQ to compress convolutional and FC layers using a clustering technique designed to minimize the reconstruction error of the layer outputs, followed by end-to-end training of cluster centroids via distillation. Their approach does not optimize the grouping of the network parameters for quantization.

Martinez \textit{et al.} \cite{martinez2021permute} used an annealed quantization algorithm to reduce quantization error further. Their procedure consists of three steps: 1) permute: search for a permutation of each layer that results in subvectors that are easier to quantize; 2) quantize: obtain codes and codebooks for each layer by minimizing the reconstruction error between approximated weights and the permuted weights; and 3) fine-tune on training dataset via gradient-based optimization, which is costly.

These vector methods quantize subvectors, depend on layer layout, and usually fine-tune. They are not a drop-in substitute for structure-agnostic scalar weight-sharing.

In summary, compressing deep neural networks while preserving their accuracy and performance by weight-sharing compression is a challenging task. In our view, the performance of the compressed network is highly dependent on two criteria: 1) the quality of grouping weights in the whole network and 2) the approximation of representative weight to be shared among weights of the network with efficiency.

\subsection{Weight Sharing as Search Space Reduction for Fine-Tuning}

Weight-sharing compression is usually motivated by storage. The same grouping also maps the parameter vector onto a much smaller decision space, namely a codebook of $K$ shared values or a blocking of coordinates. Subsequent fine-tuning can then search a space of dimension $d \ll N$ rather than the full extrinsic space.

By taking into consideration the histogram of pretrained weights, Gheorghe and Ivanovici \cite{gheorghe2021model} proposed a model-based weight quantization method that uses the double exponential probability density function to define quantization limits. However, their work is limited to modifying fully connected layers with convolutional layers and changing activation functions to implement on a hardware FPGA device.

Recently, Khosrowshahli \textit{et al.} \cite{khosrowshahli2024massive} proposed an efficient hybrid fine-tuning of DNNs approach using evolutionary metaheuristics. They argued that training deep neural networks such as ResNet-18 using gradient-descent (GD) optimization without learning-rate schedulers does not guarantee generalization. We are aware that training DNNs using GD has a quick convergence due to the nature of being a local search algorithm. In this work, they suggested a sequential hybridization of GD and gradient-free optimization. To achieve a gradient-free optimization, they employed the macro-averaged F1-score as the objective function. For parameter fine-tuning, they utilized an evolutionary metaheuristic algorithm, specifically differential evolution (DE). This algorithm uses a population of $NP$ candidate solutions to enhance diversity within the search space, necessitating a substantial DRAM capacity of $2 \times NP \times d$, where $d$ is the decision space dimension ($d < N$).

Instead of decreasing the population size $NP$, they devised a massive dimension reduction method that uses histogram uniform blocking of weights after training by GD. Previously, the efficiency of random blocking of dimensions was demonstrated on large global closed-box optimization problems where the dimension of search space is $d=10,000$ and $d=100,000$ \cite{khosrowshahli2023block}. The reduction of search space dimensionality provides opportunities for population-based evolutionary optimization algorithms like DE, PSO, and CMA-ES. In addition, multi-task learning problems with several objectives can be effectively addressed by evolutionary multi-objective algorithms such as NSGA-II \cite{deb2002fast} (for up to three objectives) and NSGA-III \cite{deb2013evolutionary} (for more than three).

To make the motivation of this work explicit, we summarize the limitations of prior work as follows. (1) many existing approaches rely on computationally expensive clustering procedures or second-order information, which limits scalability to large networks; (2) several methods optimize quantization layer by layer or through bi-level procedures, which may miss globally preferable compression solutions for the whole network; (3) previous pipelines often depend on retraining, codebooks that are specific to each layer, or other design choices that reduce practical efficiency and complicate deployment; (4) treating compression as ``how few codewords can we keep'' ignores that Huffman-coded storage depends on occupancy entropy, so a merge judged by accuracy alone can miss cheaper index streams; and (5) most prior methods treat the compression problem with assumptions that are specific to an architecture or designs that depend on the layer, rather than as a closed-box optimization problem, which limits their generalizability across diverse DNN structures, including modern architectures such as Large Language Models (LLMs) and Vision Transformers, where layer heterogeneity and scale make approaches that depend on architecture impractical. These limitations motivate our proposed framework: UBQ for low-complexity quantization, exact entropy-constrained merging of neighboring bins inside each evaluation, and closed-box multi-objective search over $(K_0,\rho)$ for the trade-off between accuracy and compression at the network level.

\begin{figure*}
    \input{figures/pipeline_diagram}
    \caption{Overview of the proposed pipeline. Flattened parameters are quantized by UBQ, merged by ECSQ, and Huffman-coded inside each NSGA-II evaluation.}
    \label{fig:diagram}
\end{figure*}

\section{Proposed Method}\label{sec:proposed_method}
Let $\theta_{\mathrm{base}} \in \mathbb{R}^N$ be the pretrained parameter vector, formed by flattening each layer's weights and then concatenating them, without using layer boundaries, tensor shapes, or channel counts. UBQ maps this vector onto $K_0$ equal-width bins in $\mathcal{O}(N)$ time, storing each occupied bin by its empirical mean and dropping empty bins. ECSQ then merges neighboring bins by an exact dynamic program on those bin statistics, and Huffman coding stores the assignment indices. These stages run inside every evaluation of a closed-box search over $(K_0,\rho)$, so the merge is scored as part of the candidate rather than applied after the search. No shared weight is retrained. We refer to this complete pipeline as UB-ECSQ. An overview of the pipeline is shown in Figure~\ref{fig:diagram}.

Algorithm~\ref{alg:alg1} gives the complete procedure, and every other algorithm in this section is one of its subroutines. Before the search starts, \textsc{FindKBounds} (Algorithm~\ref{alg:kbounds}, Section~\ref{sec:kbounds}) runs once as a warm-up and fixes the range $[K_0^{\mathrm{lb}},K_0^{\mathrm{ub}}]$ of bin counts to explore. The initial population of $NP$ individuals $x=(K_0,\rho)$ takes its $K_0$ coordinates from \textsc{LogSpacing} (Algorithm~\ref{alg:logspace}) and its $\rho$ coordinates uniformly from $[0,1]$. Each individual is scored by \textsc{EvaluateECSQ} (Algorithm~\ref{alg:fitness}). This evaluation quantizes $\theta_{\mathrm{base}}$ into $K_0$ equal-width bins with UBQ (Section~\ref{sec:ubq}) and builds the merge ladder (Algorithm~\ref{alg:ladder}, Section~\ref{sec:ladder}) by solving the exact entropy-constrained merge (Algorithm~\ref{alg:ecsq}, Section~\ref{sec:ecsq}) for each $\lambda$ on a grid. It then selects the rung closest to the target set by $\rho$ and returns the calibration error and the reciprocal of the Huffman-coded compression ratio (Section~\ref{sec:huffman}). NSGA-II evolves the population for $G$ generations with crossover, mutation, and non-dominated survival (Section~\ref{sec:mo_ecsq}), and returns the Pareto set of compressed models. The rest of this section describes these components from the bottom up, starting with the quantizer and ending with the search.

\begin{algorithm}[!t]
\caption{Overall framework: multi-objective optimization over compression decision variables with NSGA-II.}
\label{alg:alg1}

\small
\begin{algorithmic}[1]
\Require pretrained network $\mathcal{N}$ with parameters $\theta_{\mathrm{base}}$; calibration split $\mathcal{D}_c$; thresholds $\epsilon^{\min},\epsilon^{\max}$; population size $NP$; generations $G$
\Ensure non-dominated individuals after $G$ generations

\State $(K_0^{\mathrm{lb}},K_0^{\mathrm{ub}}) \gets \Call{FindKBounds}{\mathcal{N},\theta_{\mathrm{base}},\ldots}$ \Comment{Alg.~\ref{alg:kbounds}}
\State Initialize $P$ with $K_0$ from \Call{LogSpacing}{$K_0^{\mathrm{lb}},K_0^{\mathrm{ub}},NP$} and $\rho$ uniform in $[0,1]$. \Comment{Alg.~\ref{alg:logspace}}
\State $P \gets \Call{Repair}{P}$
\For{$x \in P$}
    \State $F(x) \gets \Call{EvaluateECSQ}{x}$ \Comment{Alg.~\ref{alg:fitness}}
\EndFor
\For{$g = 1, \dots, G$}
    \State $Q \gets \Call{CrossoverAndMutate}{P}$ \Comment{SBX, polynomial mutation}
    \State $Q \gets \Call{Repair}{Q}$
    \For{$x \in Q$}
        \State $F(x) \gets \Call{EvaluateECSQ}{x}$ \Comment{Alg.~\ref{alg:fitness}}
    \EndFor
    \State $P \gets \Call{SurviveNSGA2}{P \cup Q, NP}$
\EndFor
\State \Return $\{x \in P : x \text{ is non-dominated in } P\}$
\end{algorithmic}

\end{algorithm}

\subsection{Uniform Binning Quantization (UBQ)}\label{sec:ubq}
Let $\theta^{\min}$ and $\theta^{\max}$ be the minimum and maximum entries of $\theta_{\mathrm{base}}$. A scalar $\theta$ (indexed $\theta_j$) denotes an entry of that vector.

After training, the empirical histogram of the weights is typically bell-shaped and concentrated near zero \cite{han2015deep,park2017weighted,banner2019post}. A clustering fit would track that histogram more tightly. UBQ still uses equal-width cuts because they are a single pass over the $N$ weights and do not inspect architecture.
The common bin width on $[\theta^{\min}, \theta^{\max}]$ is
\begin{equation}
    \Delta = \frac{\theta^{\max} - \theta^{\min}}{K_0-1}.
\end{equation}
The grid is anchored at zero. Bin centers lie on the lattice $\{k\Delta : k \in \mathbb{Z}\}$, so zero is the center of a bin. Almost all weights lie near zero, so the bin that contains zero holds the largest share of them. Anchoring keeps that bin symmetric about zero, so its mean stays near zero for every $K_0$, instead of shifting with the position of zero inside the bin. With $k_0 = \mathrm{round}(\theta^{\min}/\Delta)$, the $i$-th bin, for $i=1,\ldots,K_0$, is the half-open interval
\begin{equation}
    \bigl[(k_0+i-\tfrac{3}{2})\Delta,\, (k_0+i-\tfrac{1}{2})\Delta\bigr).
\end{equation}
The bin index assigned to a parameter $\theta \in [\theta^{\min},\, \theta^{\max}]$ is
\begin{equation}
    b(\theta) = \mathrm{round}\!\left(\frac{\theta}{\Delta}\right) - k_0,
\end{equation}
so that $b(\theta) \in \{0, 1, \ldots, K_0-1\}$, with $\theta^{\min}$ in the first bin and $\theta^{\max}$ in the last. We denote by $\mathcal{B}_i = \{\theta_j : b(\theta_j) = i-1\}$ the set of parameters assigned to the $i$-th bin, and by $\mathcal{B} = \{\mathcal{B}_1, \ldots, \mathcal{B}_{K_0}\}$ the full collection of bins.

Determining the range of parameters can run in $\mathcal{O}(N)$ time complexity, computing bin widths involves $\mathcal{O}(K_0)$ time complexity, and assigning each parameter to the bin it belongs to runs in $\mathcal{O}(N)$ time complexity. Since $|\mathcal{B}| = K_0 < N$, the total time complexity for our proposed quantization method with $N$ weights is $\mathcal{O}(N)$.

Since the bin widths are equal and calculated based on two extreme points, min and max parameter values, it is expected to see bins that do not locate any parameters. As a result, the empty bins need to be removed, which reduces the expected uniform intervals $K_0$ to a lower and unknown number of non-empty bins $\tilde{K} \leq K_0$. These $\tilde{K}$ uniform bins $\{\mathcal{B}_i : |\mathcal{B}_i|>0\}$ are the input to entropy-constrained merging, not the stored codebook. Throughout, $\mathcal{B}$ is the collection of equal-width bins and $\mathcal{C}$ is the codebook after merging. $K_0$ is the requested bin count, $\tilde{K}$ the number of non-empty bins, and $K$ the number of codewords in $\mathcal{C}$ (same as $K$ used by K-means). Each codeword group $\mathcal{C}_i$ is a union of one or more neighboring $\mathcal{B}$'s, with occupancy $|\mathcal{C}_i|$.

Using the indices of bins and centroid values, we compute the initial centroids of the non-empty bins to map the $N$ weights to $\tilde{K}$ shared values. A centroid value $c_i$ for the $i$-th non-empty bin is calculated by non-weighted averaging on the weights contained in the bin as follows:
\begin{equation}
    c_i = \frac{1}{|\mathcal{B}_i|} \sum_{\theta_j \in \mathcal{B}_i} \theta_j, \qquad |\mathcal{B}_i|>0
\end{equation}
This is the distinction from a generic uniform grid. 
Equal-width cuts alone would reconstruct interval $i$ by its midpoint $(k_0+i-1)\Delta$, independently of which weights landed there, and would keep a reconstruction point even for an empty interval. 
UBQ uses the same cuts, but the stored value is the mean $c_i$ of the occupied bin, and empty bins are removed. 
The Uniform Quantization baseline of Choi \textit{et al.}~\cite{choi2016towards} uses this same centroid rule. It is UBQ without ECSQ merging and without the $(K_0,\rho)$ search, with $K$ swept over powers of two.

That mean is also the shared parameter we store. Pipelines such as Deep Compression \cite{han2015deep} cluster the weights and then retrain the shared centroids to recover accuracy. Retraining is a second, expensive optimization. It is an iterative gradient-based procedure on the constrained network. The training budget needed to recover accuracy is not known in advance and must be large enough to regain the baseline. The fit is prone to overfitting the training data. We do not run that stage. Each $c_i$ is computed in one pass from the weights in the bin and stored without further adjustment, with no fine-tuning of the codewords and no gradient update of the compressed model. Using a group mean as the representative follows center-based sampling \cite{rahnamayan2009center}, later applied to clustered search points by Khosrowshahli \textit{et al.} \cite{khosrowshahli4939115population, khosrowshahli2023ranking, khosrowshahli2022clustering}. Those works insert the cluster center as a new candidate in an ongoing search. We instead compute the bin mean once and store it as the shared weight.


\subsection{Entropy-Constrained Merging}\label{sec:ecsq}

After UBQ, some neighboring bins are very similar, or one holds very few weights compared with its neighbor. Merging such neighboring bins shrinks the number of codewords while adding little extra squared error, and it can make the index stream easier to compress with entropy coding.

This second effect is central to the method. When the compressed model is stored, per-weight indices are not stored as fixed-width integers since they are compressed with Huffman coding \cite{huffman1952method,han2015deep}, which assigns shorter codes to indices that occur more often. The storage cost of the index stream therefore depends on how unevenly the weights are spread across codeword groups, not merely on how many codewords remain. A codebook with many codewords but a highly skewed occupancy can compress better than a codebook with fewer codewords but a nearly uniform occupancy. The relevant design question is therefore not ``how few codewords can we keep'', but how to shape the occupancy distribution so that it is as compressible as possible for a given allowed distortion.

Entropy-Constrained Scalar Quantization (ECSQ) \cite{chou1989entropy} answers this question directly. A merge is a partition $\Pi$ of the ordered non-empty bins into contiguous groups. For a fixed $\lambda \geq 0$, we take the following:
\begin{equation}
    \Pi^*(\lambda) = \arg\min_{\Pi} J(\Pi), \qquad J(\Pi) = \mathrm{SE}(\Pi) + \lambda R(\Pi),
    \label{eq:ecsq}
\end{equation}
Where $\mathrm{SE}(\Pi)$ is the extra distortion squared error introduced by the merge, $R(\Pi)$ is the rate which is the number of bits per weight that an entropy code requires for the resulting index stream, computed from the occupancy distribution, and $\lambda$ trades one off against the other. A larger $\lambda$ favors fewer, more compressible codewords at the cost of more distortion and $\lambda=0$ leaves every non-empty bin unchanged.

\textbf{Distortion and rate from bin statistics.}
Because bins are ordered along the weight axis, the only admissible merges are of neighbors. 
Let $n_b=|\mathcal{B}_b|$ be the occupancy of non-empty bin $b$, and let $c_b$ be its centroid. With $M_0=S_0=T_0=0$, the prefix sums for $j=1,\ldots,\tilde{K}$ are as follows:
\begin{align}
    M_j &= \sum_{b=1}^{j} n_b, \nonumber \\
    S_j &= \sum_{b=1}^{j} n_b c_b, \nonumber \\
    T_j &= \sum_{b=1}^{j} n_b c_b^{2}.
    \label{eq:prefix}
\end{align}
Merging bins $i{+}1$ through $j$ into one group has occupancy $\mu=M_j-M_i$. Distortion and rate of that merge satisfy the following:
\begin{align}
    \mathrm{SE}_{i+1:j} &= (T_j - T_i) - \frac{(S_j - S_i)^{2}}{\mu}, \nonumber \\
    R_{i+1:j} &= -\mu\log_{2}(\mu/N).
    \label{eq:distortion}
\end{align}
Where $\mathrm{SE}_{i+1:j}$ is the extra squared error of merging bins $i{+}1$ through $j$ into one codeword (i.e., the distortion of merge) and $R_{i+1:j}$ is that merge's contribution to the unnormalized occupancy entropy, in bits. 
Each group $g\in\Pi$ is a contiguous block of bins $i{+}1$ through $j$. We write $\mathrm{SE}_g$ for $\mathrm{SE}_{i+1:j}$ and $R_g$ for $R_{i+1:j}$. Distortion and rate of the full partition are:
\begin{align}
    \mathrm{SE}(\Pi)&=\sum_{g\in\Pi} \mathrm{SE}_g, \nonumber \\
    R(\Pi)&=\sum_{g\in\Pi} R_g.
    \label{eq:Rpi}
\end{align}
Both $SE(\Pi)$ and $R(\Pi)$ quantities are obtained from the bin counts and centroids alone which translates to no forward pass through the network, no calibration data, and no gradients are required. 
We keep the original Shannon-rate notation $R$ of \eqref{eq:ecsq} for this contribution and it differs from the normalized bits-per-weight entropy $-p\log_2 p$ by the constant factor $N$, which is absorbed into the numerical scale of $\lambda$.

\textbf{Exact dynamic program.}
Distortion and rate are additive over contiguous groups, so for a fixed $\lambda$ the partition that minimizes \eqref{eq:ecsq} can be found exactly by dynamic programming \cite{bellman1957dynamic} rather than by trial and error. Let $dp_j$ be the minimum Lagrangian cost of partitioning the prefix $\mathcal{B}_{1},\ldots,\mathcal{B}_{j}$, with $dp_0=0$. 
The cost $dp_j$ satisfies the following recurrence:
\begin{equation}
    dp_j = \min_{0\leq i<j}\Bigl( dp_i + \mathrm{SE}_{i+1:j} + \lambda R_{i+1:j}\Bigr).
    \label{eq:dp}
\end{equation}
The minimizing index $i$ is stored as a predecessor, and the optimal partition is recovered by backtracking from $j=\tilde{K}$. Each recovered group of bins $a$ through $b$ is represented by the centroid weighted by occupancy
\begin{equation*}
    c'_{a:b} = \frac{S_b - S_{a-1}}{M_b - M_{a-1}}.
\end{equation*}
After $O(\tilde{K})$ prefix sums, each candidate interval is $O(1)$, so the recurrence runs in $O(\tilde{K}^{2})$ time with $\tilde{K}\leq K_0$. In practice this takes microseconds even for several hundred bins. On small instances we compared the dynamic program against exhaustive enumeration of contiguous partitions and obtained identical optima. Algorithm~\ref{alg:ecsq} states the procedure.

\begin{algorithm}[!t]
\caption{Entropy-constrained bin merging (exact, by dynamic programming).}
\label{alg:ecsq}

\small
\begin{algorithmic}[1]
\Require bin occupancies $n_1, \dots, n_{\tilde{K}} > 0$; bin centroids $c_1 \le \dots \le c_{\tilde{K}}$; trade-off weight $\lambda \ge 0$
\Ensure group index $\pi[1..\tilde{K}]$; merged centroids $c'_1, \dots, c'_{|\Pi|}$

\Function{MergeCost}{$M, S, T, N, \lambda, i, j$}
    \State $\mu \gets M_j - M_i$
    \State $\sigma \gets S_j - S_i$
    \State $\mathrm{SE} \gets (T_j - T_i) - \sigma^2 / \mu$
    \State $R \gets -\mu \log_2(\mu / N)$
    \State \Return $\mathrm{SE} + \lambda R$
\EndFunction

\Function{DPMerge}{$n, c, \lambda$}
    \State $N \gets \sum_{b=1}^{\tilde{K}} n_b$
    \State $M_0, S_0, T_0 \gets 0, 0, 0$ \Comment{prefix sums, 0-indexed}
    \For{$j = 1, \dots, \tilde{K}$}
        \State $M_j \gets M_{j-1} + n_j$
        \State $S_j \gets S_{j-1} + n_j c_j$
        \State $T_j \gets T_{j-1} + n_j c_j^2$
    \EndFor

    \State $dp_0 \gets 0$
    \For{$j = 1, \dots, \tilde{K}$}
        \State $dp_j \gets \infty$
        \For{$i = 0, \dots, j-1$}
            \State $\mathrm{cost} \gets \Call{MergeCost}{M, S, T, N, \lambda, i, j}$
            \If{$dp_i + \mathrm{cost} < dp_j$}
                \State $dp_j \gets dp_i + \mathrm{cost}$
                \State $\mathrm{parent}_j \gets i$
            \EndIf
        \EndFor
    \EndFor

    \State $\mathrm{groups} \gets$ empty list; $j \gets \tilde{K}$
    \While{$j > 0$} \Comment{backtrack}
        \State $i \gets \mathrm{parent}_j$
        \State prepend $(i{+}1, j)$ to $\mathrm{groups}$
        \State $j \gets i$
    \EndWhile

    \For{$g = 1, \dots, |\mathrm{groups}|$, with $(a, b) = \mathrm{groups}_g$}
        \State $\pi[a..b] \gets g$
        \State $c'_g \gets (S_b - S_{a-1}) / (M_b - M_{a-1})$
    \EndFor

    \State \Return $\pi$, $\{c'_1, \dots, c'_{|\Pi|}\}$
\EndFunction
\end{algorithmic}

\end{algorithm}

Fixing $\lambda$ and minimizing $J$ scalarizes the $\mathrm{SE}$ and $R$ trade-off into a single number and collapses it to one ordinary minimization, which Algorithm~\ref{alg:ecsq} solves exactly. This inner problem is not the multi-objective search whose Pareto front we report. Sweeping $\lambda$ over a grid, as the next subsection does to build the merge ladder, is the classical Lagrangian relaxation way of tracing a sequence of scalarized $\mathrm{SE}$ and $R$ points. The outer search in Section~\ref{sec:mo_ecsq} never scalarizes accuracy loss and $1/\mathrm{CR}$ into one number and it evolves a population of size $NP$ toward the non-dominated front of those two objectives.

\subsection{Merge Ladder}\label{sec:ladder}

To expose a one-dimensional family of merges for a given $K_0$, we solve Algorithm~\ref{alg:ecsq} once for each of a fixed increasing grid $\lambda_1<\cdots<\lambda_L$ spanning near zero (almost no merging) to comparatively large (merge almost everything). 
For every distinct number of codewords that appears across those $\lambda$, we keep the partition produced by the smallest $\lambda$ that attains it. 
The merge ladder is that table, with one row per attained codebook size and the partition kept for that size. 
Algorithm~\ref{alg:ladder} states the construction.

\begin{algorithm}[!t]
\caption{Building the merge ladder and reaching a target number of codewords $K$.}
\label{alg:ladder}

\small
\begin{algorithmic}[1]
\Require bin occupancies $n_1,\dots,n_{\tilde{K}}$; bin centroids $c_1 \le \dots \le c_{\tilde{K}}$; grid $\lambda_1 < \dots < \lambda_L$; target $K$
\Ensure the stored partition whose size is closest to $K$

\Function{BuildLadder}{$n, c, \lambda_1,\dots,\lambda_L$}
    \State $\mathrm{ladder} \gets$ empty map
    \For{$t = 1, \dots, L$}
        \State $(\pi, C') \gets$ \Call{DPMerge}{$n, c, \lambda_t$}
        \State $s \gets |C'|$
        \If{$s \notin \mathrm{ladder}$}
            \State $\mathrm{ladder}[s] \gets (\pi, C')$ \Comment{keep smallest $\lambda$ for size $s$}
        \EndIf
    \EndFor
    \State \Return $\mathrm{ladder}$
\EndFunction

\Function{NearestRung}{$\mathrm{ladder}, K$}
    \State $s^{\star} \gets \arg\min_{s \in \mathrm{keys}(\mathrm{ladder})} |s - K|$ \Comment{ties to larger $s$}
    \State \Return $s^{\star}, \mathrm{ladder}[s^{\star}]$
\EndFunction
\end{algorithmic}

\end{algorithm}

Because each call is $O(\tilde{K}^{2})$ and $L$ is small, building the ladder is negligible compared with a single forward pass on the calibration split.

The search does not request a $\lambda$. It requests a target number of codewords $K$ as a fraction $\rho\in[0,1]$ of the number of non-empty bins $\tilde{K}$ realized by UBQ:
\begin{equation}
    K = \max\bigl(1,\,\mathrm{round}(\rho\cdot \tilde{K})\bigr).
    \label{eq:rho}
\end{equation}
Thus $\rho=1$ asks for no merging and smaller $\rho$ asks for a more aggressive merge. \textsc{NearestRung} in Algorithm~\ref{alg:ladder} then returns the stored partition whose number of codewords is closest to $K$. If two stored sizes are equally close, the larger size is used. Each merged group is represented by the mean centroid weighted by occupancy from Algorithm~\ref{alg:ecsq}, and all weights in the group are replaced by that codeword. The resulting codebook $\mathcal{C}(K_0,\rho)=\{\mathcal{C}_1,\ldots,\mathcal{C}_K\}$ has $K\leq \tilde{K}\leq K_0$ codewords, with each $\mathcal{C}_i$ a contiguous merge of uniform bins.

\subsection{Automatic Search Range Selection}\label{sec:kbounds}

The bounds $K_0^{\mathrm{lb}}$ and $K_0^{\mathrm{ub}}$ that the search in Section~\ref{sec:mo_ecsq} explores are not chosen by hand per architecture. 
They are set once, before the search starts, by a warm-up that bisects accuracy vs.\ $K_0$ for UBQ alone (no merge, no pruning), so the resulting bounds reflect only how many bins this checkpoint's own weights need, independent of whatever merge or search strategy runs afterward.

Two thresholds define the two edges of the range, and they are deliberately asymmetric. 
Let the relative accuracy drop of UBQ at $K_0$ bins be
\begin{equation}
    \mathrm{Err}(K_0) = \frac{\mathrm{Acc}_{\mathrm{base}} - \mathrm{Acc}(K_0)}{\mathrm{Acc}_{\mathrm{base}}}.
    \label{eq:err}
\end{equation}
We assume that this quantity is non-increasing in $K_0$. 
The lower bound $K_0^{\mathrm{lb}}$ is the largest $K_0$ whose error is still above a loose threshold $\epsilon^{\min}=50\%$. 
This is the aggressive compression edge of the range, and it is kept on the ``still bad'' side of the accuracy cliff on purpose. A solution just past the cliff has a large error but a very large compression gain, so it can still be Pareto-optimal, and excluding it from the range would hide it from the search entirely.
The value $\epsilon^{\min}=50\%$ also keeps the range wide, which is what NSGA-II needs. Crossover and mutation are random, so the more spread out the population is over $K_0$, the more different parents the operators can combine and the better the chance that a child lands on a new part of the front. A looser threshold such as $80\%$ or $90\%$ would stretch the range even lower, but models in that extra region are damaged badly enough that almost none of their offspring survive selection, and the front we recover ends up narrower instead of wider. At $50\%$ the low edge is still aggressive, and the individuals there remain competitive enough to keep contributing to the search. 
The upper bound $K_0^{\mathrm{ub}}$ is the smallest $K_0$ whose error already drops to or below a tight threshold $\epsilon^{\max}=0.5\%$. Past this point extra bins buy no new trade-off, so the range can stop there.

Both crossings are found by bisecting $\log_2 K_0$ rather than scanning every $K_0$, and the bisection is continuous in $\log_2 K_0$ rather than stepped by whole powers of two. On a checkpoint with a sharp accuracy cliff, the true threshold crossing can fall strictly between two consecutive powers of two. Stepping only in integers would then report a bound whose actual error is nowhere near the intended threshold. The bisection converges to within a fixed resolution in $\log_2 K_0$, with a default of 0.1 or about a 7\% relative tolerance on $K_0$. This procedure requires only a handful of extra evaluations. The cost is negligible because this is a one-time warm-up rather than a cost per candidate. If the whole tested range already satisfies a threshold, or never satisfies it, that edge of the range is reported directly rather than extrapolated, with a warning in the latter case. If both thresholds collapse to the same point along a nearly vertical cliff, the upper edge is given a small amount of headroom above the lower one so the range is never empty.

Finally, both edges are snapped outward to the nearest power of two to match the encoding used for every other bin count in this work. Specifically, $K_0^{\mathrm{lb}}$ is rounded down and $K_0^{\mathrm{ub}}$ is rounded up. This direction is necessary because error is assumed to be non-increasing in $K_0$. Rounding $K_0^{\mathrm{lb}}$ down decreases $K_0$ and increases error, which can only push it further above $\epsilon^{\min}$. Rounding $K_0^{\mathrm{ub}}$ up increases $K_0$ and decreases error, which can only push it further below $\epsilon^{\max}$. Rounding in the opposite directions could cross the threshold that each bound is meant to guarantee. Algorithm~\ref{alg:kbounds} states the full procedure.

\begin{algorithm}[!t]
\caption{Automatic search range selection by threshold bisection.}
\label{alg:kbounds}

\small
\begin{algorithmic}[1]
\Function{Error}{$\mathcal{N},\, \theta_{\mathrm{base}},\, \mathrm{Acc}_{\mathrm{base}},\, \log_2 K_0$}
    \State $K_0 \leftarrow \max(1,\, \mathrm{round}(2^{\log_2 K_0}))$.
    \State $\mathcal{C} \leftarrow \Call{UniformBin}{\theta_{\mathrm{base}},\, K_0}$.
    \State \Return $\bigl(\mathrm{Acc}_{\mathrm{base}} - \mathrm{Acc}(\mathcal{N}, \mathcal{C})\bigr) / \mathrm{Acc}_{\mathrm{base}}$.
\EndFunction

\Function{Bisect}{$\mathcal{N},\, \theta_{\mathrm{base}},\, \mathrm{Acc}_{\mathrm{base}},\, \tau,\, a,\, b,\, \delta$}
    \While{$b - a > \delta$}
        \State $z \leftarrow (a+b)/2$.
        \If{$\Call{Error}{\mathcal{N},\theta_{\mathrm{base}},\mathrm{Acc}_{\mathrm{base}},z} \le \tau$}
            \State $b \leftarrow z$.
        \Else
            \State $a \leftarrow z$.
        \EndIf
    \EndWhile
    \State \Return $(a, b)$.
\EndFunction

\Function{FindKBounds}{$\mathcal{N},\, \theta_{\mathrm{base}},\, \mathrm{Acc}_{\mathrm{base}},\, \epsilon^{\min},\, \epsilon^{\max},\, [\ell, h],\, \delta$}
    \State $e_\ell \leftarrow \Call{Error}{\mathcal{N},\theta_{\mathrm{base}},\mathrm{Acc}_{\mathrm{base}},\ell}$.
    \State $e_h \leftarrow \Call{Error}{\mathcal{N},\theta_{\mathrm{base}},\mathrm{Acc}_{\mathrm{base}},h}$.
    \If{$e_\ell \le \epsilon^{\min}$}
        \State $\log_2 K_0^{\mathrm{lb}} \leftarrow \ell$.
    \ElsIf{$e_h > \epsilon^{\min}$}
        \State $\log_2 K_0^{\mathrm{lb}} \leftarrow h$. \Comment{warn}
    \Else
        \State $(\log_2 K_0^{\mathrm{lb}}, \cdot) \leftarrow \Call{Bisect}{\mathcal{N},\allowbreak\theta_{\mathrm{base}},\allowbreak\mathrm{Acc}_{\mathrm{base}},\allowbreak\epsilon^{\min},\allowbreak \ell,\allowbreak h,\allowbreak \delta}$.
    \EndIf
    \If{$e_h > \epsilon^{\max}$}
        \State $\log_2 K_0^{\mathrm{ub}} \leftarrow h$. \Comment{warn}
    \Else
        \State $(\cdot, \log_2 K_0^{\mathrm{ub}}) \leftarrow \Call{Bisect}{\mathcal{N},\allowbreak\theta_{\mathrm{base}},\allowbreak\mathrm{Acc}_{\mathrm{base}},\allowbreak\epsilon^{\max},\allowbreak \log_2 K_0^{\mathrm{lb}},\allowbreak h,\allowbreak \delta}$.
    \EndIf
    \If{$\log_2 K_0^{\mathrm{ub}} \le \log_2 K_0^{\mathrm{lb}} < h$}
        \State $\log_2 K_0^{\mathrm{ub}} \leftarrow \min(\log_2 K_0^{\mathrm{lb}} + \delta,\, h)$.
    \EndIf
    \State $\log_2 K_0^{\mathrm{lb}} \leftarrow \lfloor \log_2 K_0^{\mathrm{lb}} \rfloor$;\quad $\log_2 K_0^{\mathrm{ub}} \leftarrow \lceil \log_2 K_0^{\mathrm{ub}} \rceil$.
    \If{$\log_2 K_0^{\mathrm{ub}} \le \log_2 K_0^{\mathrm{lb}}$}
        \State $\log_2 K_0^{\mathrm{ub}} \leftarrow \log_2 K_0^{\mathrm{lb}} + 1$.
    \EndIf
    \State \Return $\bigl(2^{\log_2 K_0^{\mathrm{lb}}},\, 2^{\log_2 K_0^{\mathrm{ub}}}\bigr)$.
\EndFunction
\end{algorithmic}

\end{algorithm}

As one concrete example, this warm-up narrows ResNet-18 on CIFAR-10 to $[K_0^{\mathrm{lb}},K_0^{\mathrm{ub}}]=[64,256]$, well inside the outer range described in Section~\ref{sec:experimental_settings}.

\subsection{Multi-objective Optimization}\label{sec:mo_ecsq}

Compression with loss can be formulated as multi-objective search problem. 
We aim to preserve predictive performance while reducing the file size. So, it would be two-objective problem. The first objective is the top-1 accuracy of the compressed network on the calibration split. The second objective is the compression ratio of the stored model, which accounts for both the $K$ floating-point codewords and the Huffman-coded index stream in \eqref{eq:cr_huffman}. 

Let $x=(K_0,\rho)$, with $K_0\in[K_0^{\mathrm{lb}},K_0^{\mathrm{ub}}]$ and $\rho\in[0,1]$. We formulate the objectives as follows:
\begin{align}
    \min_{x}\; f_1(x) &= 1 - \textsc{Top1Acc}\bigl(\mathcal{N},\,\mathcal{D}_c,\,\mathcal{C}(x)\bigr), \label{eq:f1obj} \\
    \min_{x}\; f_2(x) &= \frac{1}{\mathrm{CR}_{\mathrm{H}}\bigl(\mathcal{C}(x)\bigr)}, \label{eq:f2obj}
\end{align}
Where $\mathrm{CR}_{\mathrm{H}}$ is \eqref{eq:cr_huffman} evaluated on the merged codewords. The function $\textsc{Top1Acc}$ gives the fraction of calibration examples correctly classified by network $\mathcal{N}$ compressed with codebook $\mathcal{C}(x)$. NSGA-II \cite{deb2002fast} searches $x$ and returns a Pareto set of compressed models. A user can then select a point according to the desired criterion: the maximum allowed accuracy loss, the minimum required $\mathrm{CR}$, or a knee of the front.


Each evaluation constructs $\mathcal{C}(K_0,\rho)$ through UBQ, removal of empty bins, construction of the merge ladder, and nearest-rung selection according to \eqref{eq:rho}. 
It then performs one forward pass on the calibration split for $f_1$ and one Huffman pass for $f_2$, as detailed in Algorithm~\ref{alg:fitness}.

\begin{algorithm}[!t]
\caption{Candidate solution evaluation.}
\label{alg:fitness}

\small
\begin{algorithmic}[1]
\Require individual $x = (K_0, \rho)$; pretrained model; calibration split
\Ensure objective vector $(f_1, f_2)$

\Function{EvaluateECSQ}{$x$}
    \State $(K_0, \rho) \gets x$
    \State $(n, c) \gets \Call{UniformBin}{\text{model}, K_0}$
    \State $\tilde{K} \gets |c|$
    \State $\mathrm{ladder} \gets \Call{BuildLadder}{n, c, \lambda_1,\dots,\lambda_L}$
    \State $K \gets \max(1, \mathrm{round}(\rho \cdot \tilde{K}))$
    \State $(K, \mathrm{rung}) \gets \Call{NearestRung}{\mathrm{ladder}, K}$
    \State $\mathrm{model}' \gets \Call{ApplyMerge}{\text{model}, \mathrm{rung}}$
    \State $a \gets \Call{Accuracy}{\mathrm{model}', \text{calibration split}}$
    \State $\mathrm{CR} \gets \Call{CompressionRatio}{\mathrm{model}'}$
    \State $f_1 \gets 1 - a$
    \State $f_2 \gets 1 / \mathrm{CR}$
    \State \Return $(f_1, f_2)$
\EndFunction
\end{algorithmic}

\end{algorithm}

The merge requires no forward pass and is therefore inexpensive enough to perform during every evaluation. We do not retrain the shared weights. The population is initialized by logarithmically spacing the $K_0$ coordinates over $[K_0^{\mathrm{lb}},K_0^{\mathrm{ub}}]$, as specified in Algorithm~\ref{alg:logspace}. The coordinates are evenly spaced in $\log_2 K_0$ rather than in $K_0$, consistent with the bisection in Section~\ref{sec:kbounds}. The $\rho$ coordinates are initialized uniformly over $[0,1]$.

\begin{algorithm}[!t]
\caption{Logarithmic spacing of the initial $K_0$ coordinates.}
\label{alg:logspace}

\small
\begin{algorithmic}[1]
\Function{LogSpacing}{$K_0^{\mathrm{lb}},\; K_0^{\mathrm{ub}},\; NP$}
    \State $\ell \leftarrow \log_2 K_0^{\mathrm{lb}}$;\quad $h \leftarrow \log_2 K_0^{\mathrm{ub}}$.
    \State $\gamma \leftarrow \dfrac{h - \ell}{NP - 1}$.
    \State $result \leftarrow [\,]$.
    \For{$i \gets 1$ \textbf{to} $NP$}
        \State $value \leftarrow 2^{\,\ell + (i - 1) \times \gamma}$.
        \State Append $value$ to $result$.
    \EndFor
    \State \Return $result$.
\EndFunction
\end{algorithmic}

\end{algorithm}

With these subroutines in place, the full search is Algorithm~\ref{alg:alg1}, given at the start of this section. A repair step canonicalizes each individual after initialization and after every generation. Each generation produces an offspring population $Q$ of size $NP$. After $G$ generations, non-dominated sorting of the parent population $P$ of size $NP$ yields the reported Pareto front.

\subsection{Huffman Coding}\label{sec:huffman}

The stored model consists of $K$ floating-point codewords and a per-weight index stream. The index stream dominates the file size, so the indices are Huffman-coded \cite{huffman1952method} rather than stored as fixed-width integers, following Han \textit{et al.} \cite{han2015deep} and Choi \textit{et al.} \cite{choi2016towards}. Huffman coding itself is not new in neural network compression. We use it as the lossless encoding of the assignment produced by ECSQ. Unlike a post-hoc encoding of codewords that were designed without rate in mind, Huffman length is anticipated during merging. The inner rate $R$ in \eqref{eq:ecsq} is the occupancy entropy that Huffman coding realizes. The outer objective $f_2$ is the reciprocal of the Huffman-coded compression ratio \eqref{eq:cr_huffman} on the merged codebook.

\section{Experiments}\label{sec:experiments}

\begin{table*}[t]
\centering
\resizebox{\textwidth}{!}{%
\begin{tabular}{lcccccccccccc}
\toprule
\multirow{3}{*}{Method} & \multicolumn{2}{c}{\multirow{2}{*}{$\Delta\mathrm{Acc}^{\ast}$}} & \multicolumn{10}{c}{$\Delta\mathrm{Acc}$} \\
\cmidrule(lr){4-13}
 & & & \multicolumn{2}{c}{$-0.25\%$} & \multicolumn{2}{c}{$-0.5\%$} & \multicolumn{2}{c}{$-1\%$} & \multicolumn{2}{c}{$-2\%$} & \multicolumn{2}{c}{$-5\%$} \\
\cmidrule(lr){2-3}\cmidrule(lr){4-5}\cmidrule(lr){6-7}\cmidrule(lr){8-9}\cmidrule(lr){10-11}\cmidrule(lr){12-13}
 & CR & Acc & CR & Acc & CR & Acc & CR & Acc & CR & Acc & CR & Acc \\
\midrule
\multicolumn{13}{c}{ResNet-18 ($\mathrm{Acc}_{\mathrm{base}} = 95.10\%$)} \\
\midrule
\textit{Target} & --- & $\geq 94.98$ & --- & $\geq 94.85$ & --- & $\geq 94.60$ & --- & $\geq 94.10$ & --- & $\geq 93.10$ & --- & $\geq 90.10$ \\
K-means~\cite{han2015deep} &
$5.0$ & $95.02$ & $5.0$ & $95.02$ & $6.0$ & $94.78$ & $7.4$ & $94.56$ & $9.3$ & $93.28$ & $9.3$ & $93.28$ \\
Hessian K-means~\cite{choi2016towards} &
$5.4$ & $94.98$ & $8.5$ & $94.88$ & $13.6$ & $94.78$ & $13.6$ & $94.78$ & $13.6$ & $94.78$ & $13.6$ & $94.78$ \\
Uniform Quant.~\cite{choi2016towards} &
$8.0$ & $95.04$ & $10.6$ & $94.90$ & $14.8$ & $94.84$ & $14.8$ & $94.84$ & $14.8$ & $94.84$ & $20.6$ & $92.66$ \\
Iterative ECSQ~\cite{choi2016towards} &
$3.9$ & $95.04$ & $3.9$ & $95.04$ & $3.9$ & $95.04$ & $3.9$ & $95.04$ & $3.9$ & $95.04$ & $3.9$ & $95.04$ \\
GPTQ~\cite{frantar2023gptq} &
$7.9$ & $95.00$ & $7.9$ & $95.00$ & $7.9$ & $95.00$ & $7.9$ & $95.00$ & $7.9$ & $95.00$ & $7.9$ & $95.00$ \\
Dupuis \textit{et al.}~\cite{dupuis2022heuristic} &
$12.9$ & $95.00$ & $13.5$ & $94.92$ & $14.9$ & $94.62$ & $16.4$ & $94.16$ & $17.6$ & $93.16$ & $19.9$ & $90.22$ \\
UB-ECSQ (ours) &
$\mathbf{16.2}_{\pm 0.2}$ & $94.99_{\pm 0.01}$ & $\mathbf{16.4}_{\pm 0.1}$ & $94.89_{\pm 0.05}$ & $\mathbf{20.5}_{\pm 0.0}$ & $94.68_{\pm 0.00}$ & $\mathbf{20.8}_{\pm 0.3}$ & $94.44_{\pm 0.16}$ & $\mathbf{21.5}_{\pm 0.0}$ & $93.60_{\pm 0.03}$ & $\mathbf{21.7}_{\pm 0.4}$ & $90.57_{\pm 0.62}$ \\
\bottomrule
\end{tabular}%
}
\caption{\textbf{CIFAR-10 / ResNet-18}. CR and top-1 Acc (\%) of the most compressed point meeting each column's Target. $\Delta\mathrm{Acc}$ is the Target minus $\mathrm{Acc}_{\mathrm{base}}$, the pretrained Acc. $\Delta\mathrm{Acc}^{\ast}$ is that of the highest Acc all methods reach. Each UB-ECSQ entry is the mean $\pm$ std over seeds. Bold is the highest CR.}
\label{tab:cifar10_ecsq}
\end{table*}

\begin{table*}[t]
\centering
\resizebox{\textwidth}{!}{%
\begin{tabular}{lcccccccccccc}
\toprule
\multirow{3}{*}{Method} & \multicolumn{2}{c}{\multirow{2}{*}{$\Delta\mathrm{Acc}^{\ast}$}} & \multicolumn{10}{c}{$\Delta\mathrm{Acc}$} \\
\cmidrule(lr){4-13}
 & & & \multicolumn{2}{c}{$-0.25\%$} & \multicolumn{2}{c}{$-0.5\%$} & \multicolumn{2}{c}{$-1\%$} & \multicolumn{2}{c}{$-2\%$} & \multicolumn{2}{c}{$-5\%$} \\
\cmidrule(lr){2-3}\cmidrule(lr){4-5}\cmidrule(lr){6-7}\cmidrule(lr){8-9}\cmidrule(lr){10-11}\cmidrule(lr){12-13}
 & CR & Acc & CR & Acc & CR & Acc & CR & Acc & CR & Acc & CR & Acc \\
\midrule
\multicolumn{13}{c}{ResNet-18 ($\mathrm{Acc}_{\mathrm{base}} = 76.66\%$)} \\
\midrule
\textit{Target} & --- & $\geq 76.62$ & --- & $\geq 76.41$ & --- & $\geq 76.16$ & --- & $\geq 75.66$ & --- & $\geq 74.66$ & --- & $\geq 71.66$ \\
K-means~\cite{han2015deep} &
$4.0$ & $76.68$ & $4.6$ & $76.48$ & $5.4$ & $76.22$ & $5.4$ & $76.22$ & $6.5$ & $75.08$ & $6.5$ & $75.08$ \\
Hessian K-means~\cite{choi2016towards} &
$3.9$ & $76.64$ & $5.8$ & $76.60$ & $5.8$ & $76.60$ & $12.0$ & $76.04$ & $12.0$ & $76.04$ & $14.8$ & $73.22$ \\
Uniform Quant.~\cite{choi2016towards} &
$5.3$ & $76.64$ & $8.1$ & $76.58$ & $8.1$ & $76.58$ & $10.9$ & $76.10$ & $10.9$ & $76.10$ & $10.9$ & $76.10$ \\
Iterative ECSQ~\cite{choi2016towards} &
$3.8$ & $76.62$ & $3.8$ & $76.62$ & $3.8$ & $76.62$ & $3.8$ & $76.62$ & $3.8$ & $76.62$ & $8.8$ & $74.14$ \\
GPTQ~\cite{frantar2023gptq} &
$4.0$ & $76.62$ & $5.3$ & $76.58$ & $10.6$ & $76.18$ & $10.6$ & $76.18$ & $10.6$ & $76.18$ & $10.6$ & $76.18$ \\
Dupuis \textit{et al.}~\cite{dupuis2022heuristic} &
$\mathbf{11.2}$ & $76.64$ & $\mathbf{11.3}$ & $76.48$ & $12.1$ & $76.22$ & $12.7$ & $75.66$ & $13.4$ & $74.78$ & $15.8$ & $71.72$ \\
UB-ECSQ (ours) &
$9.8_{\pm 0.0}$ & $76.72_{\pm 0.07}$ & $11.1_{\pm 0.5}$ & $76.44_{\pm 0.03}$ & $\mathbf{12.3}_{\pm 0.1}$ & $76.20_{\pm 0.03}$ & $\mathbf{12.9}_{\pm 0.2}$ & $75.85_{\pm 0.12}$ & $\mathbf{15.2}_{\pm 0.0}$ & $74.68_{\pm 0.01}$ & $\mathbf{18.0}_{\pm 0.1}$ & $71.88_{\pm 0.03}$ \\
\bottomrule
\end{tabular}%
}
\caption{\textbf{CIFAR-100 / ResNet-18}. CR and top-1 Acc (\%) of the most compressed point meeting each column's Target. $\Delta\mathrm{Acc}$ is the Target minus $\mathrm{Acc}_{\mathrm{base}}$, the pretrained Acc. $\Delta\mathrm{Acc}^{\ast}$ is that of the highest Acc all methods reach. Each UB-ECSQ entry is the mean $\pm$ std over seeds. Bold is the highest CR.}
\label{tab:cifar100_ecsq}
\end{table*}

\begin{table*}[t]
\centering
\resizebox{\textwidth}{!}{%
\begin{tabular}{lcccccccccccc}
\toprule
\multirow{3}{*}{Method} & \multicolumn{2}{c}{\multirow{2}{*}{$\Delta\mathrm{Acc}^{\ast}$}} & \multicolumn{10}{c}{$\Delta\mathrm{Acc}$} \\
\cmidrule(lr){4-13}
 & & & \multicolumn{2}{c}{$-0.25\%$} & \multicolumn{2}{c}{$-0.5\%$} & \multicolumn{2}{c}{$-1\%$} & \multicolumn{2}{c}{$-2\%$} & \multicolumn{2}{c}{$-5\%$} \\
\cmidrule(lr){2-3}\cmidrule(lr){4-5}\cmidrule(lr){6-7}\cmidrule(lr){8-9}\cmidrule(lr){10-11}\cmidrule(lr){12-13}
 & CR & Acc & CR & Acc & CR & Acc & CR & Acc & CR & Acc & CR & Acc \\
\midrule
\multicolumn{13}{c}{ResNet-18 ($\mathrm{Acc}_{\mathrm{base}} = 69.58\%$)} \\
\midrule
\textit{Target} & --- & $\geq 69.52$ & --- & $\geq 69.33$ & --- & $\geq 69.08$ & --- & $\geq 68.58$ & --- & $\geq 67.58$ & --- & $\geq 64.58$ \\
K-means~\cite{han2015deep} &
$4.7$ & $69.66$ & $4.7$ & $69.66$ & $5.6$ & $69.24$ & $5.6$ & $69.24$ & $6.6$ & $67.60$ & $6.6$ & $67.60$ \\
Hessian K-means~\cite{choi2016towards} &
$4.3$ & $69.52$ & $5.1$ & $69.46$ & $5.1$ & $69.46$ & $6.7$ & $68.70$ & $6.7$ & $68.70$ & $6.7$ & $68.70$ \\
Uniform Quant.~\cite{choi2016towards} &
$5.5$ & $69.70$ & $6.8$ & $69.40$ & $6.8$ & $69.40$ & $8.6$ & $68.96$ & $8.6$ & $68.96$ & $11.8$ & $67.56$ \\
Iterative ECSQ~\cite{choi2016towards} &
$3.6$ & $69.66$ & $3.6$ & $69.66$ & $3.6$ & $69.66$ & $3.6$ & $69.66$ & $3.6$ & $69.66$ & $3.6$ & $69.66$ \\
GPTQ~\cite{frantar2023gptq} &
$4.0$ & $69.52$ & $5.3$ & $69.46$ & $5.3$ & $69.46$ & $7.9$ & $68.90$ & $7.9$ & $68.90$ & $10.6$ & $66.60$ \\
UB-ECSQ (ours) &
$\mathbf{8.5}_{\pm 0.2}$ & $69.65_{\pm 0.05}$ & $\mathbf{8.9}_{\pm 0.1}$ & $69.40_{\pm 0.04}$ & $\mathbf{9.3}_{\pm 0.0}$ & $69.10_{\pm 0.03}$ & $\mathbf{10.5}_{\pm 0.1}$ & $68.76_{\pm 0.07}$ & $\mathbf{11.6}_{\pm 0.3}$ & $67.64_{\pm 0.04}$ & $\mathbf{13.7}_{\pm 0.1}$ & $64.91_{\pm 0.42}$ \\
\midrule
\multicolumn{13}{c}{ResNet-34 ($\mathrm{Acc}_{\mathrm{base}} = 73.46\%$)} \\
\midrule
\textit{Target} & --- & $\geq 73.38$ & --- & $\geq 73.21$ & --- & $\geq 72.96$ & --- & $\geq 72.46$ & --- & $\geq 71.46$ & --- & $\geq 68.46$ \\
K-means~\cite{han2015deep} &
$4.7$ & $73.56$ & $4.7$ & $73.56$ & $4.7$ & $73.56$ & $4.7$ & $73.56$ & $5.6$ & $72.32$ & $6.8$ & $69.62$ \\
Hessian K-means~\cite{choi2016towards} &
$4.6$ & $73.52$ & $4.6$ & $73.52$ & $5.9$ & $73.16$ & $5.9$ & $73.16$ & $9.1$ & $72.28$ & $9.1$ & $72.28$ \\
Uniform Quant.~\cite{choi2016towards} &
$\mathbf{9.0}$ & $73.46$ & $9.0$ & $73.46$ & $9.0$ & $73.46$ & $9.0$ & $73.46$ & $12.4$ & $71.62$ & $12.4$ & $71.62$ \\
Iterative ECSQ~\cite{choi2016towards} &
$3.8$ & $73.38$ & $3.8$ & $73.38$ & $3.8$ & $73.38$ & $3.8$ & $73.38$ & $3.8$ & $73.38$ & $4.5$ & $69.40$ \\
GPTQ~\cite{frantar2023gptq} &
$5.3$ & $73.52$ & $5.3$ & $73.52$ & $8.0$ & $73.08$ & $8.0$ & $73.08$ & $8.0$ & $73.08$ & $10.6$ & $71.18$ \\
UB-ECSQ (ours) &
$8.9_{\pm 0.3}$ & $73.44_{\pm 0.03}$ & $\mathbf{9.2}_{\pm 0.1}$ & $73.32_{\pm 0.08}$ & $\mathbf{9.8}_{\pm 0.0}$ & $73.01_{\pm 0.03}$ & $\mathbf{10.9}_{\pm 0.1}$ & $72.56_{\pm 0.06}$ & $\mathbf{12.9}_{\pm 0.1}$ & $71.68_{\pm 0.25}$ & $\mathbf{15.2}_{\pm 0.2}$ & $68.74_{\pm 0.31}$ \\
\midrule
\multicolumn{13}{c}{ResNet-50 ($\mathrm{Acc}_{\mathrm{base}} = 76.34\%$)} \\
\midrule
\textit{Target} & --- & $\geq 76.18$ & --- & $\geq 76.09$ & --- & $\geq 75.84$ & --- & $\geq 75.34$ & --- & $\geq 74.34$ & --- & $\geq 71.34$ \\
K-means~\cite{han2015deep} &
$3.6$ & $76.18$ & $3.6$ & $76.18$ & $4.7$ & $75.88$ & $4.7$ & $75.88$ & $5.5$ & $75.32$ & $5.5$ & $75.32$ \\
Hessian K-means~\cite{choi2016towards} &
$4.9$ & $76.32$ & $4.9$ & $76.32$ & $4.9$ & $76.32$ & $6.4$ & $75.76$ & $9.3$ & $75.00$ & $11.3$ & $72.06$ \\
Uniform Quant.~\cite{choi2016towards} &
$5.2$ & $76.20$ & $6.3$ & $76.16$ & $8.0$ & $75.94$ & $8.0$ & $75.94$ & $10.7$ & $74.90$ & $10.7$ & $74.90$ \\
Iterative ECSQ~\cite{choi2016towards} &
$4.0$ & $76.24$ & $4.0$ & $76.24$ & $4.0$ & $76.24$ & $4.0$ & $76.24$ & $4.0$ & $76.24$ & $4.0$ & $76.24$ \\
GPTQ~\cite{frantar2023gptq} &
$4.0$ & $76.26$ & $5.3$ & $76.14$ & $5.3$ & $76.14$ & $7.9$ & $75.74$ & $7.9$ & $75.74$ & $10.4$ & $73.38$ \\
UB-ECSQ (ours) &
$\mathbf{8.5}_{\pm 0.4}$ & $76.30_{\pm 0.06}$ & $\mathbf{9.3}_{\pm 0.5}$ & $76.15_{\pm 0.06}$ & $\mathbf{9.6}_{\pm 0.3}$ & $76.01_{\pm 0.06}$ & $\mathbf{10.4}_{\pm 0.1}$ & $75.47_{\pm 0.08}$ & $\mathbf{11.3}_{\pm 0.0}$ & $74.56_{\pm 0.14}$ & $\mathbf{13.8}_{\pm 0.1}$ & $71.68_{\pm 0.27}$ \\
\midrule
\multicolumn{13}{c}{ResNet-101 ($\mathrm{Acc}_{\mathrm{base}} = 77.12\%$)} \\
\midrule
\textit{Target} & --- & $\geq 77.16$ & --- & $\geq 76.87$ & --- & $\geq 76.62$ & --- & $\geq 76.12$ & --- & $\geq 75.12$ & --- & $\geq 72.12$ \\
K-means~\cite{han2015deep} &
$4.1$ & $77.16$ & $4.7$ & $76.90$ & $5.6$ & $76.82$ & $5.6$ & $76.82$ & $5.6$ & $76.82$ & $5.6$ & $76.82$ \\
Hessian K-means~\cite{choi2016towards} &
$6.3$ & $77.22$ & $6.3$ & $77.22$ & $6.3$ & $77.22$ & $9.7$ & $76.40$ & $9.7$ & $76.40$ & $12.4$ & $73.28$ \\
Uniform Quant.~\cite{choi2016towards} &
$6.8$ & $77.18$ & $8.8$ & $76.88$ & $8.8$ & $76.88$ & $8.8$ & $76.88$ & $12.0$ & $75.86$ & $12.0$ & $75.86$ \\
Iterative ECSQ~\cite{choi2016towards} &
$4.0$ & $77.18$ & $4.0$ & $77.18$ & $4.0$ & $77.18$ & $4.0$ & $77.18$ & $4.0$ & $77.18$ & $4.0$ & $77.18$ \\
GPTQ~\cite{frantar2023gptq} &
$7.9$ & $77.20$ & $7.9$ & $77.20$ & $7.9$ & $77.20$ & $7.9$ & $77.20$ & $10.4$ & $75.72$ & $10.4$ & $75.72$ \\
UB-ECSQ (ours) &
$\mathbf{9.9}_{\pm 0.0}$ & $77.24_{\pm 0.03}$ & $\mathbf{10.2}_{\pm 0.0}$ & $76.98_{\pm 0.04}$ & $\mathbf{10.9}_{\pm 0.1}$ & $76.63_{\pm 0.02}$ & $\mathbf{12.3}_{\pm 0.0}$ & $76.27_{\pm 0.12}$ & $\mathbf{12.9}_{\pm 0.0}$ & $75.35_{\pm 0.14}$ & $\mathbf{15.0}_{\pm 0.1}$ & $72.54_{\pm 0.33}$ \\
\midrule
\multicolumn{13}{c}{VGG16-BN ($\mathrm{Acc}_{\mathrm{base}} = 72.92\%$)} \\
\midrule
\textit{Target} & --- & $\geq 72.86$ & --- & $\geq 72.67$ & --- & $\geq 72.42$ & --- & $\geq 71.92$ & --- & $\geq 70.92$ & --- & $\geq 67.92$ \\
K-means~\cite{han2015deep} &
$4.2$ & $73.16$ & $4.2$ & $73.16$ & $4.2$ & $73.16$ & $4.9$ & $71.94$ & $5.8$ & $71.24$ & $5.8$ & $71.24$ \\
Hessian K-means~\cite{choi2016towards} &
$6.4$ & $72.86$ & $6.4$ & $72.86$ & $6.4$ & $72.86$ & $6.4$ & $72.86$ & $9.5$ & $71.70$ & $14.6$ & $69.34$ \\
Uniform Quant.~\cite{choi2016towards} &
$10.5$ & $72.92$ & $10.5$ & $72.92$ & $15.1$ & $72.52$ & $15.1$ & $72.52$ & $15.1$ & $72.52$ & $15.1$ & $72.52$ \\
Iterative ECSQ~\cite{choi2016towards} &
$4.1$ & $72.98$ & $4.1$ & $72.98$ & $4.1$ & $72.98$ & $4.1$ & $72.98$ & $4.1$ & $72.98$ & $4.1$ & $72.98$ \\
GPTQ~\cite{frantar2023gptq} &
$5.3$ & $72.92$ & $8.0$ & $72.82$ & $10.7$ & $72.42$ & $10.7$ & $72.42$ & $10.7$ & $72.42$ & $10.7$ & $72.42$ \\
Dupuis \textit{et al.}~\cite{dupuis2022heuristic} &
$11.1$ & $72.88$ & $11.5$ & $72.80$ & $16.6$ & $72.44$ & $16.8$ & $72.04$ & $\mathbf{20.7}$ & $71.38$ & $\mathbf{21.5}$ & $68.56$ \\
UB-ECSQ (ours) &
$\mathbf{15.4}_{\pm 0.2}$ & $72.89_{\pm 0.02}$ & $\mathbf{17.0}_{\pm 0.0}$ & $72.69_{\pm 0.02}$ & $\mathbf{17.3}_{\pm 0.1}$ & $72.50_{\pm 0.08}$ & $\mathbf{18.4}_{\pm 0.1}$ & $71.98_{\pm 0.04}$ & $19.8_{\pm 0.1}$ & $71.10_{\pm 0.12}$ & $21.3_{\pm 0.1}$ & $68.20_{\pm 0.15}$ \\
\midrule
\multicolumn{13}{c}{ViT-B/16 ($\mathrm{Acc}_{\mathrm{base}} = 80.76\%$)} \\
\midrule
\textit{Target} & --- & $\geq 80.74$ & --- & $\geq 80.51$ & --- & $\geq 80.26$ & --- & $\geq 79.76$ & --- & $\geq 78.76$ & --- & $\geq 75.76$ \\
K-means~\cite{han2015deep} &
$6.2$ & $80.74$ & $6.2$ & $80.74$ & $6.2$ & $80.74$ & $6.2$ & $80.74$ & $6.2$ & $80.74$ & $6.2$ & $80.74$ \\
Hessian K-means~\cite{choi2016towards} &
$6.9$ & $80.74$ & $6.9$ & $80.74$ & $6.9$ & $80.74$ & $8.6$ & $79.88$ & $8.6$ & $79.88$ & $8.6$ & $79.88$ \\
Uniform Quant.~\cite{choi2016towards} &
$4.6$ & $80.78$ & $6.5$ & $80.66$ & $6.5$ & $80.66$ & $8.1$ & $79.78$ & $8.1$ & $79.78$ & $8.1$ & $79.78$ \\
Iterative ECSQ~\cite{choi2016towards} &
$4.1$ & $80.74$ & $4.1$ & $80.74$ & $4.1$ & $80.74$ & $4.1$ & $80.74$ & $4.1$ & $80.74$ & $4.1$ & $80.74$ \\
UB-ECSQ (ours) &
$\mathbf{7.1}_{\pm 0.1}$ & $80.77_{\pm 0.02}$ & $\mathbf{7.6}_{\pm 0.1}$ & $80.53_{\pm 0.01}$ & $\mathbf{7.8}_{\pm 0.1}$ & $80.41_{\pm 0.08}$ & $\mathbf{9.0}_{\pm 0.0}$ & $79.77_{\pm 0.01}$ & $\mathbf{9.2}_{\pm 0.0}$ & $79.04_{\pm 0.18}$ & $\mathbf{10.0}_{\pm 0.0}$ & $75.91_{\pm 0.12}$ \\
\bottomrule
\end{tabular}%
}
\caption{\textbf{ImageNet-1K}, all six architectures. CR and top-1 Acc (\%) of the most compressed point meeting each column's Target. $\Delta\mathrm{Acc}$ is the Target minus $\mathrm{Acc}_{\mathrm{base}}$, the pretrained Acc. $\Delta\mathrm{Acc}^{\ast}$ is that of the highest Acc all methods reach. Each UB-ECSQ entry is the mean $\pm$ std over seeds. Bold is the highest CR.}
\label{tab:imagenet_ecsq}
\end{table*}

\begin{table*}[t]
\centering
\resizebox{\textwidth}{!}{%
\begin{tabular}{lcccccccccccc}
\toprule
\multirow{2}{*}{Method} & \multicolumn{2}{c}{Target 1} & \multicolumn{2}{c}{Target 2} & \multicolumn{2}{c}{Target 3} & \multicolumn{2}{c}{Target 4} & \multicolumn{2}{c}{Target 5} & \multicolumn{2}{c}{Target 6} \\
\cmidrule(lr){2-3}\cmidrule(lr){4-5}\cmidrule(lr){6-7}\cmidrule(lr){8-9}\cmidrule(lr){10-11}\cmidrule(lr){12-13}
 & CR & Acc & CR & Acc & CR & Acc & CR & Acc & CR & Acc & CR & Acc \\
\midrule
\multicolumn{13}{c}{ResNet-18 ($\mathrm{Acc}_{\mathrm{base}} = 69.58\%$)} \\
\midrule
\textit{Target} & --- & $\geq 69.08$ ($-0.50\%$) & --- & $\geq 68.83$ ($-0.75\%$) & --- & $\geq 68.58$ ($-1.00\%$) & --- & $\geq 68.08$ ($-1.50\%$) & --- & $\geq 67.08$ ($-2.50\%$) & --- & $\geq 64.08$ ($-5.50\%$) \\
Dupuis \textit{et al.}~\cite{dupuis2022heuristic} &
$7.0$ & $69.08$ & $8.1$ & $68.86$ & $8.1$ & $68.86$ & $8.6$ & $68.30$ & $9.2$ & $67.20$ & $10.5$ & $64.34$ \\
UB-ECSQ (ours) &
$\mathbf{9.3}_{\pm 0.0}$ & $69.10_{\pm 0.03}$ & $\mathbf{9.8}_{\pm 0.3}$ & $68.89_{\pm 0.06}$ & $\mathbf{10.5}_{\pm 0.1}$ & $68.76_{\pm 0.07}$ & $\mathbf{10.9}_{\pm 0.1}$ & $68.23_{\pm 0.04}$ & $\mathbf{11.9}_{\pm 0.1}$ & $67.22_{\pm 0.15}$ & $\mathbf{13.8}_{\pm 0.1}$ & $64.40_{\pm 0.29}$ \\
\midrule
\multicolumn{13}{c}{ResNet-34 ($\mathrm{Acc}_{\mathrm{base}} = 73.46\%$)} \\
\midrule
\textit{Target} & --- & $\geq 72.56$ ($-0.90\%$) & --- & $\geq 72.31$ ($-1.15\%$) & --- & $\geq 72.06$ ($-1.40\%$) & --- & $\geq 71.56$ ($-1.90\%$) & --- & $\geq 70.56$ ($-2.90\%$) & --- & $\geq 67.56$ ($-5.90\%$) \\
Dupuis \textit{et al.}~\cite{dupuis2022heuristic} &
$8.7$ & $72.56$ & $8.7$ & $72.56$ & $8.7$ & $72.56$ & $8.7$ & $72.56$ & $9.7$ & $70.58$ & $10.8$ & $67.58$ \\
UB-ECSQ (ours) &
$\mathbf{10.8}_{\pm 0.1}$ & $72.64_{\pm 0.07}$ & $\mathbf{11.2}_{\pm 0.5}$ & $72.43_{\pm 0.08}$ & $\mathbf{12.2}_{\pm 0.4}$ & $72.15_{\pm 0.02}$ & $\mathbf{12.8}_{\pm 0.1}$ & $71.81_{\pm 0.24}$ & $\mathbf{13.8}_{\pm 0.1}$ & $70.62_{\pm 0.05}$ & $\mathbf{15.5}_{\pm 0.1}$ & $68.08_{\pm 0.35}$ \\
\midrule
\multicolumn{13}{c}{ResNet-50 ($\mathrm{Acc}_{\mathrm{base}} = 76.34\%$)} \\
\midrule
\textit{Target} & --- & $\geq 74.00$ ($-2.34\%$) & --- & $\geq 73.75$ ($-2.59\%$) & --- & $\geq 73.50$ ($-2.84\%$) & --- & $\geq 73.00$ ($-3.34\%$) & --- & $\geq 72.00$ ($-4.34\%$) & --- & $\geq 69.00$ ($-7.34\%$) \\
Dupuis \textit{et al.}~\cite{dupuis2022heuristic} &
$6.9$ & $74.00$ & $6.9$ & $73.98$ & $7.6$ & $73.54$ & $8.7$ & $73.00$ & $9.1$ & $72.72$ & $10.7$ & $69.32$ \\
UB-ECSQ (ours) &
$\mathbf{11.5}_{\pm 0.1}$ & $74.18_{\pm 0.08}$ & $\mathbf{12.2}_{\pm 0.3}$ & $73.85_{\pm 0.04}$ & $\mathbf{12.5}_{\pm 0.1}$ & $73.62_{\pm 0.11}$ & $\mathbf{12.7}_{\pm 0.0}$ & $73.14_{\pm 0.09}$ & $\mathbf{13.5}_{\pm 0.2}$ & $72.23_{\pm 0.18}$ & $\mathbf{14.7}_{\pm 0.0}$ & $69.37_{\pm 0.27}$ \\
\midrule
\multicolumn{13}{c}{ResNet-101 ($\mathrm{Acc}_{\mathrm{base}} = 77.12\%$)} \\
\midrule
\textit{Target} & --- & $\geq 72.08$ ($-5.04\%$) & --- & $\geq 71.83$ ($-5.29\%$) & --- & $\geq 71.58$ ($-5.54\%$) & --- & $\geq 71.08$ ($-6.04\%$) & --- & $\geq 70.08$ ($-7.04\%$) & --- & $\geq 67.08$ ($-10.04\%$) \\
Dupuis \textit{et al.}~\cite{dupuis2022heuristic} &
$8.3$ & $72.08$ & $8.3$ & $72.08$ & $8.3$ & $72.08$ & $9.0$ & $71.22$ & $9.3$ & $70.22$ & $10.1$ & $68.26$ \\
UB-ECSQ (ours) &
$\mathbf{15.0}_{\pm 0.1}$ & $72.54_{\pm 0.33}$ & $\mathbf{15.2}_{\pm 0.1}$ & $71.96_{\pm 0.13}$ & $\mathbf{15.3}_{\pm 0.1}$ & $71.77_{\pm 0.09}$ & $\mathbf{15.5}_{\pm 0.0}$ & $71.16_{\pm 0.12}$ & $\mathbf{15.6}_{\pm 0.0}$ & $70.28_{\pm 0.19}$ & $\mathbf{17.1}_{\pm 0.0}$ & $67.46_{\pm 0.10}$ \\
\midrule
\multicolumn{13}{c}{VGG16-BN ($\mathrm{Acc}_{\mathrm{base}} = 72.92\%$)} \\
\midrule
\textit{Target} & --- & $\geq 72.88$ ($-0.04\%$) & --- & $\geq 72.63$ ($-0.29\%$) & --- & $\geq 72.38$ ($-0.54\%$) & --- & $\geq 71.88$ ($-1.04\%$) & --- & $\geq 70.88$ ($-2.04\%$) & --- & $\geq 67.88$ ($-5.04\%$) \\
Dupuis \textit{et al.}~\cite{dupuis2022heuristic} &
$11.1$ & $72.88$ & $16.5$ & $72.64$ & $16.6$ & $72.44$ & $16.8$ & $72.04$ & $\mathbf{20.7}$ & $71.38$ & $\mathbf{21.5}$ & $68.56$ \\
UB-ECSQ (ours) &
$\mathbf{15.4}_{\pm 0.2}$ & $72.90_{\pm 0.02}$ & $\mathbf{17.0}_{\pm 0.0}$ & $72.67_{\pm 0.01}$ & $\mathbf{17.3}_{\pm 0.0}$ & $72.42_{\pm 0.04}$ & $\mathbf{18.4}_{\pm 0.1}$ & $71.98_{\pm 0.04}$ & $19.9_{\pm 0.0}$ & $70.93_{\pm 0.04}$ & $21.3_{\pm 0.0}$ & $68.01_{\pm 0.12}$ \\
\bottomrule
\end{tabular}%
}
\caption{\textbf{ImageNet-1K} comparison with Dupuis \textit{et al.}~\cite{dupuis2022heuristic}. CR and top-1 Acc (\%) of the most compressed point meeting each column's Target. Target 1 is the highest Acc both methods reach. Targets 2--6 lie 0.25--5\% below it. The value in parentheses is the Target minus $\mathrm{Acc}_{\mathrm{base}}$, the pretrained Acc. Each UB-ECSQ entry is the mean $\pm$ std over seeds. Bold is the highest CR.}
\label{tab:imagenet_dupuis}
\end{table*}

\begin{table*}[t]
\centering
\resizebox{\textwidth}{!}{%
\begin{tabular}{lcccccccc}
\toprule
 & CIFAR-10 & CIFAR-100 & \multicolumn{6}{c}{ImageNet-1K} \\
\cmidrule(lr){2-2}\cmidrule(lr){3-3}\cmidrule(lr){4-9}
Method & ResNet-18 & ResNet-18 & ResNet-18 & ResNet-34 & ResNet-50 & ResNet-101 & VGG16-BN & ViT-B/16 \\
\midrule
K-means~\cite{han2015deep} & $0.8706$ & $0.6766$ & $0.6107$ & $0.6360$ & $0.6554$ & $0.6646$ & $0.6322$ & $0.6985$ \\
Hessian K-means~\cite{choi2016towards} & $0.9123$ & $0.7237$ & $0.6368$ & $0.6773$ & $0.6967$ & $0.7184$ & $0.6771$ & $0.7171$ \\
Uniform Quant.~\cite{choi2016towards} & $0.9101$ & $0.7179$ & $0.6498$ & $0.6894$ & $0.7093$ & $0.7230$ & $0.6946$ & $0.7353$ \\
Iterative ECSQ~\cite{choi2016towards} & $0.7302$ & $0.7059$ & $0.5842$ & $0.6264$ & $0.6125$ & $0.6333$ & $0.6076$ & $0.6761$ \\
GPTQ~\cite{frantar2023gptq} & $0.8305$ & $0.7157$ & $0.6390$ & $0.6805$ & $0.6964$ & $0.7021$ & $0.6796$ & -- \\
Dupuis \textit{et al.}~\cite{dupuis2022heuristic} & $0.9122$ & $0.7256$ & $0.6301$ & $0.6658$ & $0.6774$ & $0.6588$ & $0.6984$ & -- \\
UB-ECSQ (ours) & $\mathbf{0.9169}_{\pm 0.0003}$ & $\mathbf{0.7358}_{\pm 0.0004}$ & $\mathbf{0.6595}_{\pm 0.0003}$ & $\mathbf{0.6992}_{\pm 0.0003}$ & $\mathbf{0.7208}_{\pm 0.0006}$ & $\mathbf{0.7331}_{\pm 0.0006}$ & $\mathbf{0.7004}_{\pm 0.0002}$ & $\mathbf{0.7475}_{\pm 0.0001}$ \\
\bottomrule
\end{tabular}%
}
\caption{Hypervolume of each method's final Pareto front (top-1 error vs.\ $1/\mathrm{CR}$, reference point $(1,1)$). Higher is better. Each UB-ECSQ entry is the mean $\pm$ std over seeds. A dash is a method that was not run. Bold is the highest per column.}
\label{tab:hypervolume}
\end{table*}

We evaluate the proposed method on CIFAR-10, CIFAR-100, and ImageNet-1K. Section~\ref{sec:experimental_settings} describes the datasets, models, and search settings. Section~\ref{sec:results} compares against previous weight-sharing quantization methods and analyzes the Pareto fronts.

\subsection{Experimental Settings} \label{sec:experimental_settings}
\textbf{Datasets.} The experiments use three public RGB image classification datasets, namely, CIFAR-10, CIFAR-100, and ImageNet-1K. CIFAR-10 and CIFAR-100 \cite{krizhevsky2009learning} have 50,000 training and 10,000 test $32 \times 32$ images in 10 and 100 classes, respectively.

Large Scale Visual Recognition Challenge (ILSVRC) ImageNet-1K \cite{deng2009imagenet} has 1.2 million training images and 50,000 validation images of varying resolution in 1000 classes.

Every candidate is scored on a calibration split that is disjoint from the network's training data and from the final test set. On CIFAR-10/100, it is a class-balanced $10\%$ of the training set ($5{,}000$ images) that is held out when the networks are trained, and the $10{,}000$ test images are kept for final testing. On ImageNet-1K, the $50{,}000$ validation images are split into a $5{,}000$-image calibration split and a disjoint $40{,}000$-image test split.

\textbf{Network Structures and Parameters.}
We adopt ResNet-[18,34,50,101] \cite{he2016deep}, VGG16-BN \cite{simonyan2014very}, and ViT-B/16 \cite{dosovitskiy2020image}. For ImageNet-1K, we use the pretrained PyTorch model-zoo checkpoints \cite{marcel2010torchvision, paszke2019pytorch}.

On CIFAR-10/100, the last layer must have 10 or 100 outputs. Those models are trained from scratch. For a fair comparison, we train ResNet-18 only. We use SGD \cite{sutskever2013importance} with multi-step learning-rate decay. The initial learning rate is $0.1$, momentum is $0.9$, and weight decay is $5 \times 10^{-4}$. The learning rate is multiplied by $0.1$ at epochs $91$ and $136$, and we train for $182$ epochs.

The CIFAR networks are trained with batch size $128$. All searches run on a single NVIDIA A100 GPU and evaluate candidates with batch size $512$ on CIFAR-10/100 and $256$ on ImageNet-1K. The proposed method does not retrain after quantization. The calibration split is used only to evaluate $f_1$ for each NSGA-II candidate. Entropy-constrained merging uses bin occupancy counts. It does not use calibration labels, and there is no merge stage after the search.

\textbf{Multi-objective Optimization.}
We use NSGA-II \cite{deb2002fast} as implemented in pymoo \cite{pymoo}, with the following settings. The population size is $NP=100$. Crossover is Simulated Binary Crossover (SBX) \cite{deb2007self} with probability $p_c=0.9$ and distribution index $\eta_c=15$. Mutation is Polynomial Mutation (PM) with probability $p_m=0.9$ and distribution index $\eta_m=10$.

The $K_0$ bounds are not set by hand. The warm-up of Section~\ref{sec:kbounds} bisects $\log_2 K_0$ inside $[2^{1},2^{10}]$ on CIFAR-10/100 and $[2^{1},2^{13}]$ on ImageNet-1K. These outer ranges let the search avoid collapsed networks at either edge. We run $G=10$ generations on CIFAR-10/100 and $G=15$ on ImageNet-1K. Each evaluation builds $\mathcal{C}(K_0,\rho)$ as in Section~\ref{sec:mo_ecsq} and records the top-1 error on the calibration split and $1/\mathrm{CR}_{\mathrm{H}}$.

\textbf{Baselines.} We reproduce four quantization schemes from Choi \textit{et al.} \cite{choi2016towards} on the same pretrained checkpoints, with no pruning and no fine-tuning. All compression ratios use Huffman-coded storage accounting, as in the UB-ECSQ rows. K-means, Hessian K-means, and Uniform Quantization are swept over $K=2^{b}$. Iterative ECSQ is swept over the Lagrange multiplier $\lambda$. In every table cell we report the most compressed feasible point that still meets that column's accuracy budget.

K-means is a global codebook in the style of Deep Compression \cite{han2015deep}, shared across layers, with K-means++ initialization, ten restarts, and one deterministic fit per $K$. Hessian K-means follows Choi \textit{et al.} \cite{choi2016towards} and uses the same global assignment with Hessian-weighted means. Their diagonal Hessian recursion \cite{lecun1987modeles,becker1988improving} has no published implementation. We use the empirical Fisher diagonal estimated from $200$ calibration images, or $500$ on CIFAR-10. This targets the same quantity as their Section~3.5 Adam second-moment alternative \cite{kingma2014adam}. Uniform Quantization uses uniformly spaced thresholds with per-bin mean centroids \cite{choi2016towards}. The reconstruction rule is the same as UBQ, with empirical means rather than midpoints. There is no ECSQ merging and no $(K_0,\rho)$ search, and $K$ is restricted to powers of two. Iterative ECSQ follows Algorithm~1 of Choi \textit{et al.} \cite{choi2016towards}, Appendix~A.3. It runs Hessian-weighted, entropy Lagrangian Lloyd iteration, initialized at $K=1024$ from that architecture's Hessian K-means solution. Setting $\lambda=0$ reduces to Hessian K-means at $K=1024$. We sweep $\lambda$ from $0$ until the realized number of codewords bottoms out.

We also compare against GPTQ \cite{frantar2023gptq}, a post-training quantization method. It rounds each layer's weights to a fixed-width integer grid, one input column at a time, and updates the not-yet-quantized weights to compensate the layer's output error, using second-order statistics of that layer's inputs on calibration data. We quantize every convolutional and fully connected layer with asymmetric scales per output channel and no grouping, calibrated on four batches of the calibration split, that is, $2{,}048$ images on CIFAR-10/100 and $1{,}024$ on ImageNet-1K. GPTQ is run at 2, 3, 4, 6, and 8 bits on CIFAR-100 and on the ImageNet-1K convolutional networks, and at 4 bits on CIFAR-10. It stores fixed-width integer codes with a 16-bit scale and zero point per output channel, so its CR is computed from that storage without Huffman coding. Unlike UB-ECSQ, it relies on layer boundaries and on calibration forward passes.

We additionally evaluate the layer-wise weight-sharing approach of Dupuis \textit{et al.}~\cite{dupuis2022heuristic} on ResNet-18 for CIFAR-10/100 and on ResNet-18/34/50/101 and VGG16-BN for ImageNet-1K. Its candidates specify a separate codebook size for each weight tensor, unlike our global codebook. We select operating points using the recorded calibration accuracy and Huffman-coded CR with the same rule as for the other methods. On CIFAR-10/100 and on ImageNet-1K VGG16-BN it is compared at the same accuracy targets in Tables \ref{tab:cifar10_ecsq}--\ref{tab:imagenet_ecsq}. On the ImageNet-1K ResNets its evaluated fronts do not come within $0.25\%$ of the pretrained accuracy, so Table \ref{tab:imagenet_dupuis} compares it with UB-ECSQ at accuracy levels that both fronts reach, and includes VGG16-BN on the same basis. One evaluated front is available per configuration, so no across run standard deviation is reported for this baseline.

\subsection{Results} \label{sec:results}

For each accuracy threshold, the tables report the highest CR among evaluated points whose calibration top-1 accuracy is at least the threshold, together with that point's accuracy. CR is Huffman-coded for every method except GPTQ, whose fixed-width codes are counted as stored. We write $\Delta\mathrm{Acc}$ for a threshold minus the pretrained accuracy $\mathrm{Acc}_{\mathrm{base}}$, in absolute percentage points of top-1 accuracy rather than relative percentages. For example, $\Delta\mathrm{Acc}$~$=-1\%$ from $95.10\%$ is $94.10\%$. The Target row gives each column's threshold, and the block title gives $\mathrm{Acc}_{\mathrm{base}}$. The first column, $\Delta\mathrm{Acc}^{\ast}$, is the $\Delta\mathrm{Acc}$ of the highest accuracy that every method in the block attains, with UB-ECSQ represented by its least accurate seed, so every method has a feasible point at every threshold. In Tables \ref{tab:cifar10_ecsq}--\ref{tab:imagenet_ecsq}, $|\Delta\mathrm{Acc}^{\ast}|\leq 0.16\%$ in every setting, so it serves as the near-lossless threshold, and the other columns are $\Delta\mathrm{Acc}$~$=-0.25\%$, $-0.5\%$, $-1\%$, $-2\%$, and $-5\%$. Table \ref{tab:imagenet_dupuis} numbers its thresholds Target~1--6. Target~1 is the highest accuracy that Dupuis \textit{et al.}\ and every UB-ECSQ seed reach, Targets~2--6 lie $0.25\%$--$5\%$ below it, and its Target row also gives each threshold's drop from the pretrained accuracy.

\begin{figure*}[t]
\centering
\includegraphics[width=0.85\textwidth]{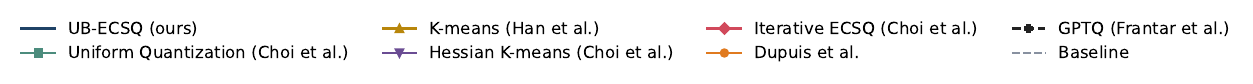}
\par\vspace{0.3em}
\begin{subfigure}{0.325\textwidth}
  \includegraphics[width=\linewidth]{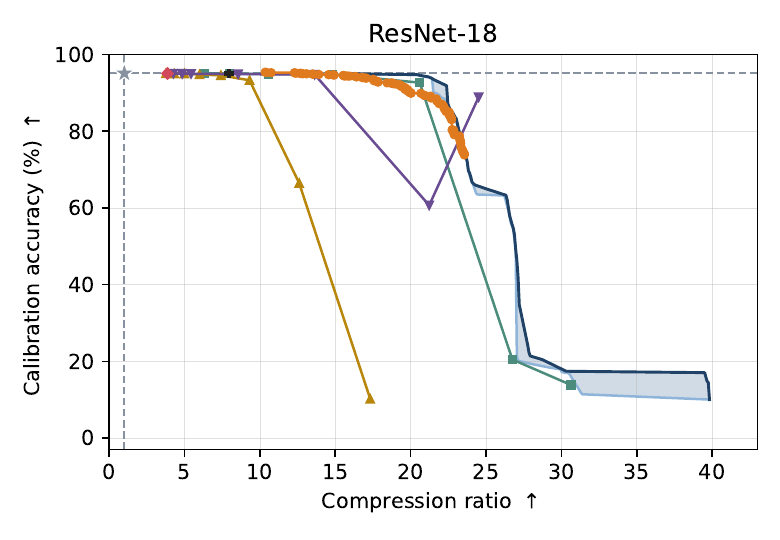}
  \caption{CIFAR-10, ResNet-18}
\end{subfigure}\hfill
\begin{subfigure}{0.325\textwidth}
  \includegraphics[width=\linewidth]{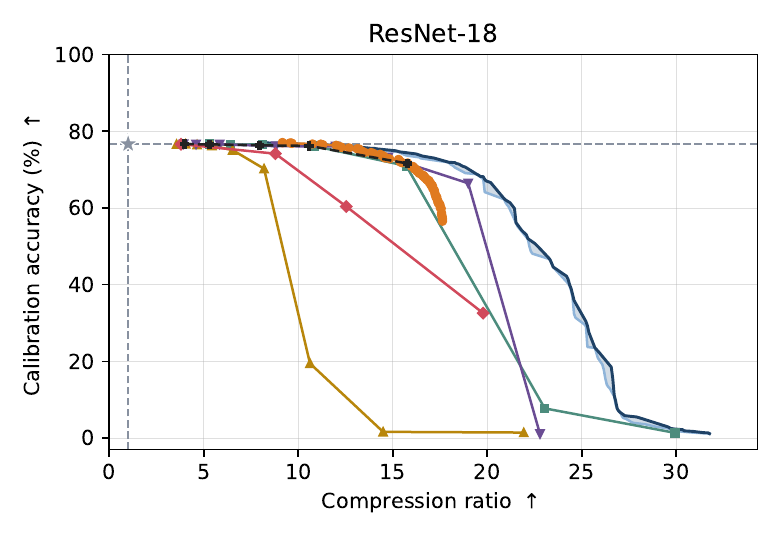}
  \caption{CIFAR-100, ResNet-18}
\end{subfigure}\hfill
\begin{subfigure}{0.325\textwidth}
  \includegraphics[width=\linewidth]{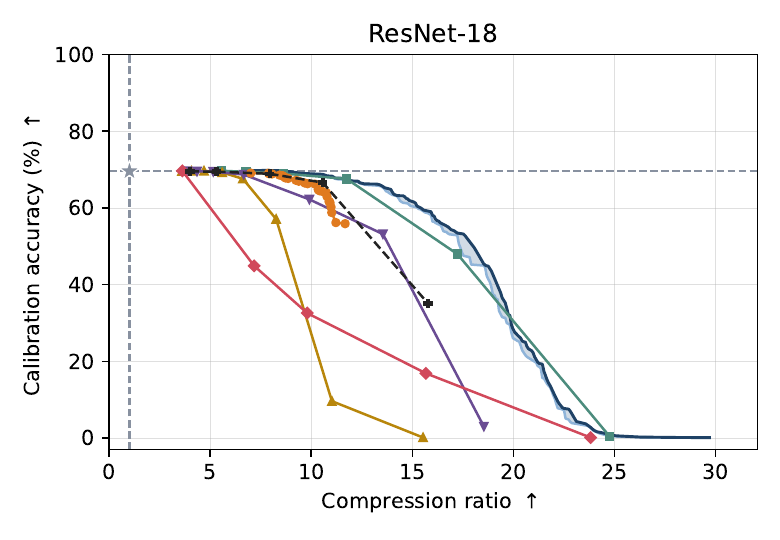}
  \caption{ImageNet, ResNet-18}
\end{subfigure}

\vspace{0.2em}

\begin{subfigure}{0.325\textwidth}
  \includegraphics[width=\linewidth]{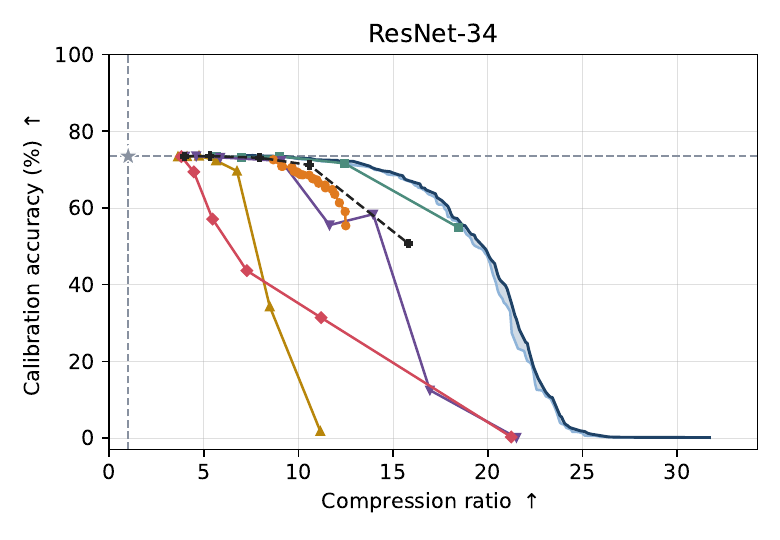}
  \caption{ImageNet, ResNet-34}
\end{subfigure}\hfill
\begin{subfigure}{0.325\textwidth}
  \includegraphics[width=\linewidth]{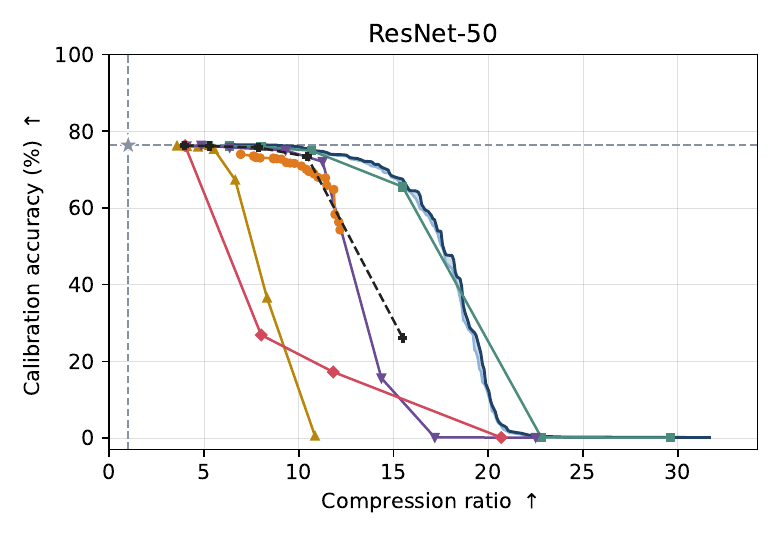}
  \caption{ImageNet, ResNet-50}
\end{subfigure}\hfill
\begin{subfigure}{0.325\textwidth}
  \includegraphics[width=\linewidth]{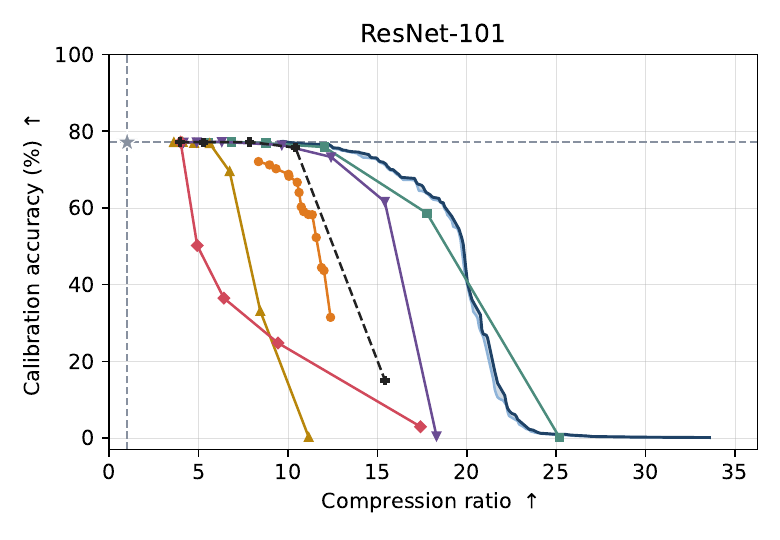}
  \caption{ImageNet, ResNet-101}
\end{subfigure}

\vspace{0.2em}

\begin{subfigure}{0.325\textwidth}
  \includegraphics[width=\linewidth]{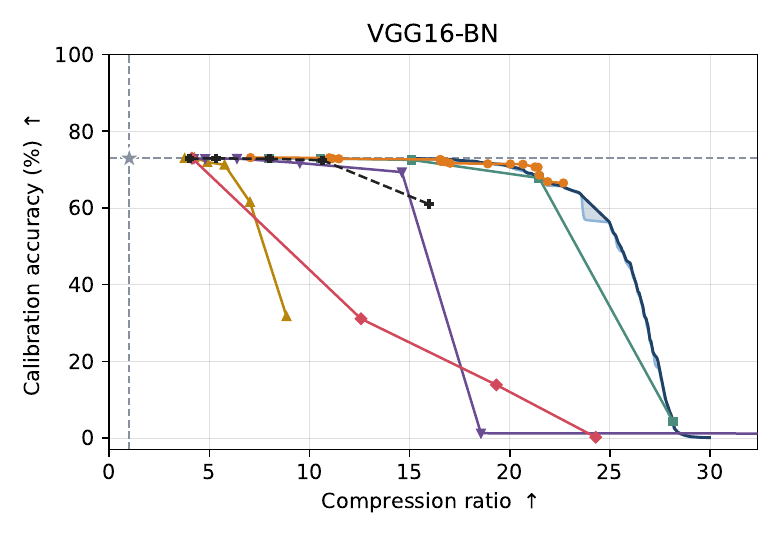}
  \caption{ImageNet, VGG16-BN}
\end{subfigure}\hspace{0.03\textwidth}
\begin{subfigure}{0.325\textwidth}
  \includegraphics[width=\linewidth]{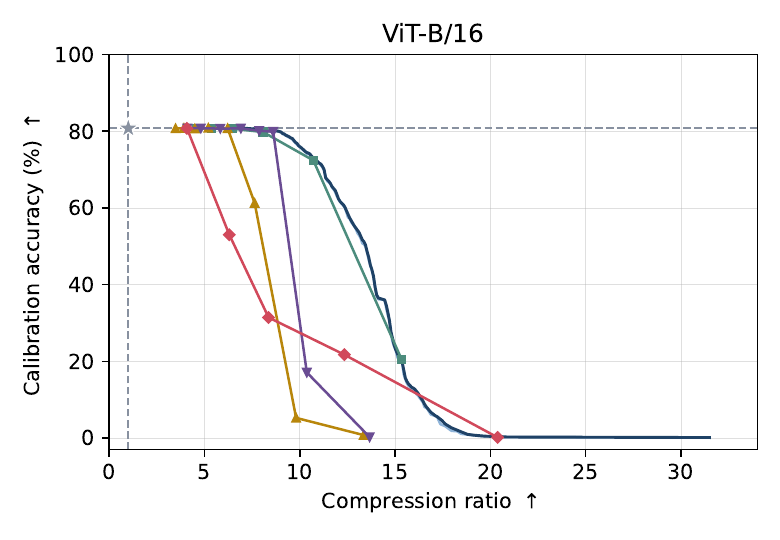}
  \caption{ImageNet, ViT-B/16}
\end{subfigure}
\caption{Pareto fronts of UB-ECSQ and all compared methods, \textbf{ResNet-18} on \textbf{CIFAR-10/100} (a,b) and \textbf{ImageNet-1K} by architecture (c--h). Each UB-ECSQ seed front is interpolated onto a common CR grid. The solid line is the best accuracy across seeds at each CR, and the shaded band is the min-to-max range. Baseline curves connect every evaluated point in order of CR, excluding points at chance accuracy.}
\label{fig:pareto_all}
\end{figure*}

\textbf{CIFAR-10/100.}
Tables \ref{tab:cifar10_ecsq} and \ref{tab:cifar100_ecsq} compare ResNet-18 on CIFAR-10 and CIFAR-100. At a $0.5\%$ accuracy drop, UB-ECSQ reaches CR $20.5$ on CIFAR-10 versus $14.9$ for Dupuis \textit{et al.}, the strongest baseline at this budget. On CIFAR-100, the same threshold gives CR $12.3$ versus $12.1$. Dupuis \textit{et al.}\ achieves higher CR at the two tightest CIFAR-100 thresholds, $\Delta\mathrm{Acc}^{\ast}$ at $76.62\%$ and $\Delta\mathrm{Acc}$~$=-0.25\%$. Its CRs there are $11.2$ and $11.3$, versus UB-ECSQ's $9.8$ and $11.1$. Figure \ref{fig:pareto_all} shows the Pareto fronts of all methods.

On CIFAR-10, the $\Delta\mathrm{Acc}^{\ast}$ threshold of $94.98\%$ is set by the least accurate UB-ECSQ seed. At this threshold UB-ECSQ reaches CR $16.2$, versus $12.9$ for Dupuis \textit{et al.}\ and at most $8.0$ for the other baselines. Above this threshold, only Dupuis \textit{et al.}\ retains the pretrained accuracy of $95.10\%$, at CR $12.6$, and its front peaks at $95.38\%$ at CR $10.4$. No UB-ECSQ seed exceeds $95.00\%$. Iterative ECSQ does not produce a family of trade-offs on this dataset. Every $\lambda>0$ collapsed to chance accuracy. That row is a single $K=1024$ point repeated across columns. The empirical Fisher diagonal is heavy-tailed. A global $\lambda$ large enough to merge high-Hessian weights already overwhelms the many low-Hessian weights. This is a property of applying their procedure with this Hessian proxy rather than a modification of it.

On CIFAR-100, UB-ECSQ reaches a best accuracy of $76.96\%$ at CR $8.6$. Iterative ECSQ has a short region of gradual degradation before the same collapse.


\textbf{ImageNet-1K.}
Table \ref{tab:imagenet_ecsq} repeats the comparison on ResNet-$\{18,34,50,101\}$, VGG16-BN, and ViT-B/16. Across all 48 targeted accuracies in Tables \ref{tab:cifar10_ecsq}--\ref{tab:imagenet_ecsq}, UB-ECSQ attains the highest CR in 43. The exceptions are CIFAR-100 at $\Delta\mathrm{Acc}^{\ast}$ and $\Delta\mathrm{Acc}$~$=-0.25\%$, where Dupuis \textit{et al.}\ leads as noted above; ImageNet/ResNet-34 at $\Delta\mathrm{Acc}^{\ast}$, $73.38\%$, where Uniform Quantization's CR $9.0$ exceeds UB-ECSQ's $8.9$; and ImageNet/VGG16-BN at $\Delta\mathrm{Acc}$~$=-2\%$ and $-5\%$, where Dupuis \textit{et al.}\ reaches CR $20.7$ and $21.5$ versus UB-ECSQ's $19.8$ and $21.3$. At the four tighter VGG16-BN thresholds UB-ECSQ leads Dupuis \textit{et al.}, for example with CR $18.4$ versus $16.8$ at $\Delta\mathrm{Acc}$~$=-1\%$. GPTQ does not attain the highest CR at any target. At a $1\%$ drop it reaches CR $7.9$--$10.7$ on the five ImageNet-1K convolutional networks, versus $10.4$--$18.4$ for UB-ECSQ. Its $2$-bit points, the most compressed it produces, retain between $15.00\%$ and $60.98\%$ accuracy on those networks.

Table \ref{tab:imagenet_dupuis} compares UB-ECSQ with Dupuis \textit{et al.}\ on the four ImageNet ResNets and on VGG16-BN. On the ResNets, the most accurate points on the evaluated Dupuis fronts lie $0.50\%$, $0.90\%$, $2.34\%$, and $5.04\%$ below the pretrained accuracy for ResNet-18, -34, -50, and -101, respectively, so each block's Target~1 is that level. On ResNet-101 the targets therefore lie $5.04\%$--$10.04\%$ below the pretrained accuracy. On VGG16-BN, Dupuis \textit{et al.}\ exceeds the pretrained accuracy, so Target~1, $72.88\%$, is set by the least accurate UB-ECSQ seed. UB-ECSQ attains the higher CR in 28 of the $30$ cells, by a factor of $1.03$ on VGG16-BN at Target~2 to $1.84$ on ResNet-101 at Target~3, and in all $24$ ResNet cells by a factor of at least $1.21$. The exceptions are VGG16-BN at Targets~5 and~6, where Dupuis \textit{et al.}\ reaches CR $20.7$ and $21.5$ versus UB-ECSQ's $19.9$ and $21.3$. At Target~1, for example, UB-ECSQ reaches CR $9.3$ versus $7.0$ on ResNet-18 and $15.0$ versus $8.3$ on ResNet-101. The accuracy offsets describe the evaluated Dupuis fronts, not limits on what that method could achieve with other search settings.

As on CIFAR, Iterative ECSQ is dominated by its $\lambda=0$ point, $K=1024$, at the tighter targets. Only ResNet-34 admits one further non-collapsed $\lambda>0$ point within the $5\%$ drop. Every architecture collapses to chance at larger $\lambda$.

\textbf{Robustness Across Seeds.}
UB-ECSQ entries summarize five independent search seeds, and every seed contributes to every cell. Baseline entries are single runs. Across Tables \ref{tab:cifar10_ecsq}--\ref{tab:imagenet_dupuis}, the seed standard deviation of UB-ECSQ is at most $0.5$ in CR and at most $0.62\%$ in accuracy. The reported standard deviations measure variation in the search on a fixed pretrained checkpoint. They do not measure variation across independently trained networks. The UB-ECSQ band in Figure \ref{fig:pareto_all} shows the min-to-max accuracy at each CR after interpolation onto a common grid. Because Dupuis \textit{et al.}\ has one evaluated front per configuration, these results do not establish statistical significance for differences between the methods.



\textbf{Effect of Binning, Merging, and Search.}
Two curves in Figure \ref{fig:pareto_all} isolate parts of the pipeline. The Uniform Quantization curve is UBQ alone, with equal-width bins, mean centroids, and empty bins dropped. There is no search and no ECSQ merge, and $K_0$ is swept over powers of two from $2$ to $2048$. The gap between this curve and the UB-ECSQ front is therefore the combined gain of entropy-constrained merging and the $(K_0,\rho)$ search.

K-means, the Deep Compression baseline of Tables \ref{tab:cifar10_ecsq}--\ref{tab:imagenet_ecsq}, is swept over the same grid. At matched accuracy, K-means needs fewer codewords than UBQ, but its index histogram is flatter. After Huffman coding it attains a lower CR for the same accuracy budget.


\textbf{Hypervolume of the Pareto Fronts.}
Tables \ref{tab:cifar10_ecsq}--\ref{tab:imagenet_dupuis} compare individual operating points. Table \ref{tab:hypervolume} summarizes each method's entire final front by its hypervolume, the area of objective space that the front dominates. The objectives are top-1 error and $1/\mathrm{CR}$, both minimized, as in the search, with CR computed as in the tables. The reference point is $(1,1)$, which corresponds to $100\%$ error at $\mathrm{CR}=1$, the uncompressed size. It is the worst value of each objective, it is the same for every method, and every evaluated point of every method dominates it, so no solution is discarded. The hypervolume therefore lies in $[0,1]$, and higher is better. UB-ECSQ hypervolumes are computed per seed and reported as mean and standard deviation. The baselines are single runs. Values are comparable within a column, not across columns, because each network has its own accuracy range.

UB-ECSQ attains the highest hypervolume in all eight settings. Its margin over the next best method ranges from $0.0020$ on VGG16-BN, over Dupuis \textit{et al.}, to $0.0122$ on ViT-B/16, over Uniform Quantization. The next best method is Dupuis \textit{et al.}\ on CIFAR-100 and VGG16-BN, Hessian K-means on CIFAR-10, and Uniform Quantization on the other five ImageNet-1K architectures. Each margin is at least $10$ times the seed standard deviation of UB-ECSQ, which is at most $0.0006$. Hence the UB-ECSQ front dominates a larger region of the trade-off space than any baseline front. This holds even where a baseline attains a higher CR at a single accuracy target in Tables \ref{tab:cifar100_ecsq} and \ref{tab:imagenet_ecsq}.

\section{Conclusion}\label{sec:conclusion}

In this paper, we proposed UB-ECSQ, a retraining- and gradient-free compression framework based on weight-sharing for pretrained deep networks. All parameters in a network are treated as one vector in which the quantization does not inspect layer structures or tensor shapes, and the shared weights are not fine-tuned after quantization. The framework separates the inexpensive construction of candidate codebooks from the evaluation of their predictive effect. Uniform binning and entropy-constrained merging build each codebook from bin statistics alone, with distortion and occupancy entropy guiding the merge, while NSGA-II scores each candidate by its single forward pass accuracy and Huffman-coded storage. Because the cost of construction is negligible next to one forward pass, every candidate compressed model is evaluated on the network itself rather than through a proxy of its accuracy.

Entropy-constrained merging attains a higher CR at a $1\%$ drop in accuracy budget than the baseline methods in all eight settings. At this budget it also exceeds the layer-wise compression method of Dupuis \textit{et al.}\ on CIFAR-10/100 and VGG16-BN, and on the ImageNet-1K ResNets at every evaluated accuracy threshold that both methods reach. Over the whole objective space, our proposed framework's Pareto front has the highest hypervolume in all eight settings. A single search yields a set of operating points, from which a practitioner can select a model that meets a memory or accuracy constraint.

Two observations extend beyond these rankings. First, when indices are Huffman-coded, the number of codewords is not a reliable measure of storage cost. At matched accuracy, K-means needs fewer codewords than UBQ, yet it attains a lower CR because its occupancy histogram is flatter. The same consideration applies to integer codes stored at fixed width, as in our GPTQ baseline, which do not exploit a skewed occupancy. A codebook intended for storage should therefore be designed against the rate of its index stream, as our proposed framework does, rather than against codebook size or bit width alone. Second, inexpensive binning is an effective foundation for jointly optimizing compression and accuracy. Without any merging or search, equal-width bins with mean centroids already give a larger hypervolume than K-means in all eight settings and than Hessian K-means on all six ImageNet-1K networks.

The main limitation of the framework is that its savings are in storage and transmission rather than in computation. The stored model must be Huffman-decoded and its codewords looked up before inference, so, unlike integer quantizers such as GPTQ, it does not by itself reduce arithmetic cost. Further limitations are that scoring each candidate requires a labeled calibration split and one forward pass, and that the comparison with Dupuis \textit{et al.}\ rests on one evaluated front per configuration.

Future work is motivated first by the variation across architectures. Within a $1\%$ drop, CR varied more across the evaluated architecture families, from $9.0$ on ViT-B/16 to $18.4$ on VGG16-BN, than across the evaluated ResNet depths, from $10.4$ to $12.3$. Because the UBQ bin width is set by the extreme weights, we conjecture that large magnitude outliers contribute to the lower CR on ViT-B/16. Clipping or separately storing such weights before binning would test this conjecture and is a natural first step toward language models, where such outliers are common. Other future work includes a small number of codebooks for groups of layers, chosen by the same closed-box search, to target the operating points where the layer-wise method leads, composition with pruning, and comparison with quantization-aware training.

\FloatBarrier
\bibliographystyle{IEEEtranS}
\bibliography{references_ieee}

@IEEEtranBSTCTL{IEEEexample:BSTcontrol,
  CTLdash_repeated_names = "no"
}

@inproceedings{banner2019post,
  author={Banner, Ron and Nahshan, Yury and Soudry, Daniel},
  title={Post training 4-bit quantization of convolutional networks for rapid-deployment},
  booktitle={Proc. Adv. Neural Inf. Process. Syst.},
  volume={32},
  pages={7948--7956},
  year={2019}
}

@inproceedings{becker1988improving,
  author={Becker, Sue and LeCun, Yann},
  title={Improving the convergence of back-propagation learning with second order methods},
  booktitle={Proc. Connectionist Models Summer School},
  pages={29--37},
  year={1988}
}

@book{bellman1957dynamic,
  author={Bellman, Richard},
  title={Dynamic Programming},
  publisher={Princeton Univ. Press},
  address={Princeton, NJ, USA},
  year={1957}
}

@article{pymoo,
  author={Blank, Julian and Deb, Kalyanmoy},
  title={pymoo: Multi-objective optimization in {Python}},
  journal={IEEE Access},
  volume={8},
  pages={89497--89509},
  year={2020}
}

@article{caruana1997multitask,
  author={Caruana, Rich},
  title={Multitask learning},
  journal={Mach. Learn.},
  volume={28},
  number={1},
  pages={41--75},
  month=jul,
  year={1997}
}

@inproceedings{choi2016towards,
  author={Choi, Yoojin and El-Khamy, Mostafa and Lee, Jungwon},
  title={Towards the limit of network quantization},
  booktitle={Proc. Int. Conf. Learn. Represent.},
  year={2017},
  note={arXiv:1612.01543}
}

@article{choi2020universal,
  author={Choi, Yoojin and El-Khamy, Mostafa and Lee, Jungwon},
  title={Universal deep neural network compression},
  journal={IEEE J. Sel. Topics Signal Process.},
  volume={14},
  number={4},
  pages={715--726},
  month=may,
  year={2020}
}

@article{chou1989entropy,
  author={Chou, Philip A and Lookabaugh, Tom and Gray, Robert M},
  title={Entropy-constrained vector quantization},
  journal={IEEE Trans. Acoust., Speech, Signal Process.},
  volume={37},
  number={1},
  pages={31--42},
  month=jan,
  year={1989}
}

@inproceedings{chou2018merging,
  author={Chou, Yi-Min and Chan, Yi-Ming and Lee, Jia-Hong and Chiu, Chih-Yi and Chen, Chu-Song},
  title={Merging deep neural networks for mobile devices},
  booktitle={Proc. IEEE Conf. Comput. Vis. Pattern Recognit. Workshops},
  pages={1686--1694},
  year={2018}
}

@article{deb2013evolutionary,
  author={Deb, Kalyanmoy and Jain, Himanshu},
  title={An evolutionary many-objective optimization algorithm using reference-point-based nondominated sorting approach, part {I}: Solving problems with box constraints},
  journal={IEEE Trans. Evol. Comput.},
  volume={18},
  number={4},
  pages={577--601},
  month=aug,
  year={2014}
}

@article{deb2002fast,
  author={Deb, Kalyanmoy and Pratap, Amrit and Agarwal, Sameer and Meyarivan, TAMT},
  title={A fast and elitist multiobjective genetic algorithm: {NSGA-II}},
  journal={IEEE Trans. Evol. Comput.},
  volume={6},
  number={2},
  pages={182--197},
  month=apr,
  year={2002}
}

@inproceedings{deb2007self,
  author={Deb, Kalyanmoy and Sindhya, Karthik and Okabe, Tatsuya},
  title={Self-adaptive simulated binary crossover for real-parameter optimization},
  booktitle={Proc. 9th Annu. Conf. Genet. Evol. Comput.},
  pages={1187--1194},
  year={2007}
}

@inproceedings{deng2009imagenet,
  author={Deng, Jia and Dong, Wei and Socher, Richard and Li, Li-Jia and Li, Kai and Fei-Fei, Li},
  title={{ImageNet}: A large-scale hierarchical image database},
  booktitle={Proc. IEEE Conf. Comput. Vis. Pattern Recognit.},
  pages={248--255},
  year={2009}
}

@inproceedings{dosovitskiy2020image,
  author={Dosovitskiy, Alexey and Beyer, Lucas and Kolesnikov, Alexander and Weissenborn, Dirk and Zhai, Xiaohua and Unterthiner, Thomas and Dehghani, Mostafa and Minderer, Matthias and Heigold, Georg and Gelly, Sylvain and others},
  title={An image is worth 16x16 words: Transformers for image recognition at scale},
  booktitle={Proc. Int. Conf. Learn. Represent.},
  year={2021},
  note={arXiv:2010.11929}
}

@inproceedings{duong2015low,
  author={Duong, Long and Cohn, Trevor and Bird, Steven and Cook, Paul},
  title={Low resource dependency parsing: Cross-lingual parameter sharing in a neural network parser},
  booktitle={Proc. 53rd Annu. Meeting Assoc. Comput. Linguistics 7th Int. Joint Conf. Natural Lang. Process.},
  pages={845--850},
  year={2015}
}

@article{dupuis2021cnn,
  author={Dupuis, Etienne and Novo, David and O'Connor, Ian and Bosio, Alberto},
  title={{CNN} weight sharing based on a fast accuracy estimation metric},
  journal={Microelectron. Rel.},
  volume={122},
  month=jul,
  year={2021},
  note={{Art. no. 114148}}
}

@inproceedings{dupuis2022heuristic,
  author={Dupuis, Etienne and Novo, David and O'Connor, Ian and Bosio, Alberto},
  title={A heuristic exploration of retraining-free weight-sharing for {CNN} compression},
  booktitle={Proc. 27th Asia South Pacific Design Autom. Conf. (ASP-DAC)},
  pages={134--139},
  year={2022}
}

@inproceedings{frantar2023gptq,
  author={Frantar, Elias and Ashkboos, Saleh and Hoefler, Torsten and Alistarh, Dan},
  title={{GPTQ}: Accurate post-training quantization for generative pre-trained transformers},
  booktitle={Proc. Int. Conf. Learn. Represent.},
  year={2023},
  note={arXiv:2210.17323}
}

@inproceedings{gheorghe2021model,
  author={Gheorghe, Stefan and Ivanovici, Mihai},
  title={Model-based weight quantization for convolutional neural network compression},
  booktitle={Proc. 16th Int. Conf. Eng. Modern Elect. Syst. (EMES)},
  pages={1--4},
  year={2021}
}

@article{gish1968asymptotically,
  author={Gish, Herbert and Pierce, John},
  title={Asymptotically efficient quantizing},
  journal={IEEE Trans. Inf. Theory},
  volume={14},
  number={5},
  pages={676--683},
  month=sep,
  year={1968}
}

@inproceedings{goddard2024arcee,
  author={Goddard, Charles and Siriwardhana, Shamane and Ehghaghi, Malikeh and Meyers, Luke and Karpukhin, Vlad and Benedict, Brian and McQuade, Mark and Solawetz, Jacob},
  title={Arcee's {MergeKit}: A toolkit for merging large language models},
  booktitle={Proc. Conf. Empirical Methods Natural Lang. Process.: Ind. Track},
  pages={477--485},
  year={2024}
}

@misc{gong2014compressing,
  author={Gong, Yunchao and Liu, Liu and Yang, Ming and Bourdev, Lubomir},
  title={Compressing deep convolutional networks using vector quantization},
  year={2014},
  note={arXiv:1412.6115}
}

@inproceedings{han2015deep,
  author={Han, Song and Mao, Huizi and Dally, William J},
  title={Deep compression: Compressing deep neural networks with pruning, trained quantization and {Huffman} coding},
  booktitle={Proc. Int. Conf. Learn. Represent.},
  year={2016},
  note={arXiv:1510.00149}
}

@inproceedings{han2015learning,
  author={Han, Song and Pool, Jeff and Tran, John and Dally, William},
  title={Learning both weights and connections for efficient neural networks},
  booktitle={Proc. Adv. Neural Inf. Process. Syst.},
  volume={28},
  pages={1135--1143},
  year={2015}
}

@inproceedings{he2015delving,
  author={He, Kaiming and Zhang, Xiangyu and Ren, Shaoqing and Sun, Jian},
  title={Delving deep into rectifiers: Surpassing human-level performance on {ImageNet} classification},
  booktitle={Proc. IEEE Int. Conf. Comput. Vis.},
  pages={1026--1034},
  year={2015}
}

@inproceedings{he2016deep,
  author={He, Kaiming and Zhang, Xiangyu and Ren, Shaoqing and Sun, Jian},
  title={Deep residual learning for image recognition},
  booktitle={Proc. IEEE Conf. Comput. Vis. Pattern Recognit.},
  pages={770--778},
  year={2016}
}

@article{huffman1952method,
  author={Huffman, David A},
  title={A method for the construction of minimum-redundancy codes},
  journal={Proc. IRE},
  volume={40},
  number={9},
  pages={1098--1101},
  month=sep,
  year={1952}
}

@inproceedings{jacob2018quantization,
  author={Jacob, Benoit and Kligys, Skirmantas and Chen, Bo and Zhu, Menglong and Tang, Matthew and Howard, Andrew and Adam, Hartwig and Kalenichenko, Dmitry},
  title={Quantization and training of neural networks for efficient integer-arithmetic-only inference},
  booktitle={Proc. IEEE Conf. Comput. Vis. Pattern Recognit.},
  pages={2704--2713},
  year={2018}
}

@inproceedings{khosrowshahli2023block,
  author={Khosrowshahli, Rasa and Rahnamayan, Shahryar},
  title={Block differential evolution},
  booktitle={Proc. IEEE Congr. Evol. Comput. (CEC)},
  pages={1--8},
  year={2023}
}

@inproceedings{khosrowshahli2022clustering,
  author={Khosrowshahli, Rasa and Rahnamayan, Shahryar and Bidgoli, Azam Asilian},
  title={Clustering center-based differential evolution},
  booktitle={Proc. IEEE Congr. Evol. Comput. (CEC)},
  pages={1--8},
  year={2022}
}

@article{khosrowshahli4939115population,
  author={Khosrowshahli, Rasa and Rahnamayan, Shahryar and Ibrahim, Amin and Asilian Bidgoli, Azam and Makrehchi, Masoud},
  title={Population-level center-based sampling for meta-heuristic algorithms},
  journal={Swarm Evol. Comput.},
  volume={92},
  month=feb,
  year={2025},
  note={{Art. no. 101827}}
}

@inproceedings{khosrowshahli2023ranking,
  author={Khosrowshahli, Rasa and Rahnamayan, Shahryar and Ibrahim, Amin and Makrehchi, Masoud},
  title={Ranking center-based {NSGA-II}},
  booktitle={Proc. IEEE Int. Conf. Syst., Man, Cybern. (SMC)},
  pages={4162--4169},
  year={2023}
}

@inproceedings{khosrowshahli2024massive,
  author={Khosrowshahli, Rasa and Rahnamayan, Shahryar and Ombuki-Berman, Beatrice},
  title={Massive dimensions reduction and hybridization with meta-heuristics in deep learning},
  booktitle={Proc. IEEE Can. Conf. Elect. Comput. Eng. (CCECE)},
  pages={469--475},
  year={2024}
}

@inproceedings{kingma2014adam,
  author={Kingma, Diederik P and Ba, Jimmy},
  title={Adam: A method for stochastic optimization},
  booktitle={Proc. Int. Conf. Learn. Represent.},
  year={2015},
  note={arXiv:1412.6980}
}

@misc{krishnamoorthi2018quantizing,
  author={Krishnamoorthi, Raghuraman},
  title={Quantizing deep convolutional networks for efficient inference: A whitepaper},
  year={2018},
  note={arXiv:1806.08342}
}

@techreport{krizhevsky2009learning,
  author={Krizhevsky, Alex and Hinton, Geoffrey},
  title={Learning multiple layers of features from tiny images},
  institution={Univ. Toronto},
  address={Toronto, ON, Canada},
  year={2009}
}

@inproceedings{krizhevsky2012imagenet,
  author={Krizhevsky, Alex and Sutskever, Ilya and Hinton, Geoffrey E},
  title={{ImageNet} classification with deep convolutional neural networks},
  booktitle={Proc. Adv. Neural Inf. Process. Syst.},
  volume={25},
  pages={1097--1105},
  year={2012}
}

@phdthesis{lecun1987modeles,
  author={LeCun, Yann},
  title={Mod{\`e}les connexionnistes de l'apprentissage},
  school={Univ. Pierre et Marie Curie},
  address={Paris, France},
  year={1987}
}

@inproceedings{li2018measuring,
  author={Li, Chunyuan and Farkhoor, Heerad and Liu, Rosanne and Yosinski, Jason},
  title={Measuring the intrinsic dimension of objective landscapes},
  booktitle={Proc. Int. Conf. Learn. Represent.},
  year={2018},
  note={arXiv:1804.08838}
}

@article{liang2021pruning,
  author={Liang, Tailin and Glossner, John and Wang, Lei and Shi, Shaobo and Zhang, Xiaotong},
  title={Pruning and quantization for deep neural network acceleration: A survey},
  journal={Neurocomputing},
  volume={461},
  pages={370--403},
  month=oct,
  year={2021}
}

@inproceedings{lin2020mcunet,
  author={Lin, Ji and Chen, Wei-Ming and Lin, Yujun and Cohn, John and Gan, Chuang and Han, Song},
  title={{MCUNet}: Tiny deep learning on {IoT} devices},
  booktitle={Proc. Adv. Neural Inf. Process. Syst.},
  volume={33},
  pages={11711--11722},
  year={2020}
}

@article{lloyd1982least,
  author={Lloyd, Stuart},
  title={Least squares quantization in {PCM}},
  journal={IEEE Trans. Inf. Theory},
  volume={28},
  number={2},
  pages={129--137},
  month=mar,
  year={1982}
}

@inproceedings{marcel2010torchvision,
  author={Marcel, S{\'e}bastien and Rodriguez, Yann},
  title={Torchvision the machine-vision package of torch},
  booktitle={Proc. 18th ACM Int. Conf. Multimedia},
  pages={1485--1488},
  year={2010}
}

@inproceedings{martinez2021permute,
  author={Martinez, Julieta and Shewakramani, Jashan and Liu, Ting Wei and B{\^a}rsan, Ioan Andrei and Zeng, Wenyuan and Urtasun, Raquel},
  title={Permute, quantize, and fine-tune: Efficient compression of neural networks},
  booktitle={Proc. IEEE/CVF Conf. Comput. Vis. Pattern Recognit.},
  pages={15699--15708},
  year={2021}
}

@article{Simplifying,
  author={Nowlan, Steven J. and Hinton, Geoffrey E.},
  title={Simplifying neural networks by soft weight-sharing},
  journal={Neural Comput.},
  volume={4},
  number={4},
  pages={473--493},
  month=jul,
  year={1992}
}

@article{ota2017deep,
  author={Ota, Kaoru and Dao, Minh Son and Mezaris, Vasileios and Natale, Francesco GB De},
  title={Deep learning for mobile multimedia: A survey},
  journal={ACM Trans. Multimedia Comput., Commun., Appl.},
  volume={13},
  number={3s},
  pages={1--22},
  month=aug,
  year={2017}
}

@inproceedings{park2017weighted,
  author={Park, Eunhyeok and Ahn, Junwhan and Yoo, Sungjoo},
  title={Weighted-entropy-based quantization for deep neural networks},
  booktitle={Proc. IEEE Conf. Comput. Vis. Pattern Recognit.},
  pages={5456--5464},
  year={2017}
}

@inproceedings{paszke2019pytorch,
  author={Paszke, Adam and Gross, Sam and Massa, Francisco and Lerer, Adam and Bradbury, James and Chanan, Gregory and Killeen, Trevor and Lin, Zeming and Gimelshein, Natalia and Antiga, Luca and others},
  title={{PyTorch}: An imperative style, high-performance deep learning library},
  booktitle={Proc. Adv. Neural Inf. Process. Syst.},
  volume={32},
  pages={8024--8035},
  year={2019}
}

@incollection{pikoulis2022new,
  author={Pikoulis, Erion Vasilis and Mavrokefalidis, Christos and Nousias, Stavros and Lalos, Aris S},
  title={A new clustering-based technique for the acceleration of deep convolutional networks},
  booktitle={Deep Learning Applications, Volume 3},
  publisher={Springer},
  address={Singapore},
  pages={123--150},
  year={2022}
}

@inproceedings{rahnamayan2009center,
  author={Rahnamayan, Shahryar and Wang, G Gary},
  title={Center-based sampling for population-based algorithms},
  booktitle={Proc. IEEE Congr. Evol. Comput.},
  pages={933--938},
  year={2009}
}

@article{rokh2023comprehensive,
  author={Rokh, Babak and Azarpeyvand, Ali and Khanteymoori, Alireza},
  title={A comprehensive survey on model quantization for deep neural networks in image classification},
  journal={ACM Trans. Intell. Syst. Technol.},
  volume={14},
  number={6},
  pages={1--50},
  month=dec,
  year={2023}
}

@misc{ruder2017overview,
  author={Ruder, Sebastian},
  title={An overview of multi-task learning in deep neural networks},
  year={2017},
  note={arXiv:1706.05098}
}

@inproceedings{simonyan2014very,
  author={Simonyan, Karen and Zisserman, Andrew},
  title={Very deep convolutional networks for large-scale image recognition},
  booktitle={Proc. Int. Conf. Learn. Represent.},
  year={2015},
  note={arXiv:1409.1556}
}

@inproceedings{son2018clustering,
  author={Son, Sanghyun and Nah, Seungjun and Lee, Kyoung Mu},
  title={Clustering convolutional kernels to compress deep neural networks},
  booktitle={Proc. Eur. Conf. Comput. Vis. (ECCV)},
  pages={216--232},
  year={2018}
}

@inproceedings{stock2019and,
  author={Stock, Pierre and Joulin, Armand and Gribonval, R{\'e}mi and Graham, Benjamin and J{\'e}gou, Herv{\'e}},
  title={And the bit goes down: Revisiting the quantization of neural networks},
  booktitle={Proc. Int. Conf. Learn. Represent.},
  year={2020},
  note={arXiv:1907.05686}
}

@inproceedings{sutskever2013importance,
  author={Sutskever, Ilya and Martens, James and Dahl, George and Hinton, Geoffrey},
  title={On the importance of initialization and momentum in deep learning},
  booktitle={Proc. Int. Conf. Mach. Learn.},
  pages={1139--1147},
  year={2013}
}

@article{sze2017efficient,
  author={Sze, Vivienne and Chen, Yu-Hsin and Yang, Tien-Ju and Emer, Joel S},
  title={Efficient processing of deep neural networks: A tutorial and survey},
  journal={Proc. IEEE},
  volume={105},
  number={12},
  pages={2295--2329},
  month=dec,
  year={2017}
}

@inproceedings{tung2018clip,
  author={Tung, Frederick and Mori, Greg},
  title={{CLIP-Q}: Deep network compression learning by in-parallel pruning-quantization},
  booktitle={Proc. IEEE Conf. Comput. Vis. Pattern Recognit.},
  pages={7873--7882},
  year={2018}
}

@inproceedings{ullrich2017soft,
  author={Ullrich, Karen and Meeds, Edward and Welling, Max},
  title={Soft weight-sharing for neural network compression},
  booktitle={Proc. Int. Conf. Learn. Represent.},
  year={2017},
  note={arXiv:1702.04008}
}

@inproceedings{wang2019haq,
  author={Wang, Kuan and Liu, Zhijian and Lin, Yujun and Lin, Ji and Han, Song},
  title={{HAQ}: Hardware-aware automated quantization with mixed precision},
  booktitle={Proc. IEEE/CVF Conf. Comput. Vis. Pattern Recognit.},
  pages={8612--8620},
  year={2019}
}

@inproceedings{wu2016quantized,
  author={Wu, Jiaxiang and Leng, Cong and Wang, Yuhang and Hu, Qinghao and Cheng, Jian},
  title={Quantized convolutional neural networks for mobile devices},
  booktitle={Proc. IEEE Conf. Comput. Vis. Pattern Recognit.},
  pages={4820--4828},
  year={2016}
}

@inproceedings{wu2018deep,
  author={Wu, Junru and Wang, Yue and Wu, Zhenyu and Wang, Zhangyang and Veeraraghavan, Ashok and Lin, Yingyan},
  title={Deep k-means: Re-training and parameter sharing with harder cluster assignments for compressing deep convolutions},
  booktitle={Proc. Int. Conf. Mach. Learn.},
  pages={5363--5372},
  year={2018}
}

@misc{yang2020dp,
  author={Yang, Dingcheng and Yu, Wenjian and Zhou, Ao and Mu, Haoyuan and Yao, Gary and Wang, Xiaoyi},
  title={{DP-Net}: Dynamic programming guided deep neural network compression},
  year={2020},
  note={arXiv:2003.09615}
}

\end{document}